%% file: root.tex
\documentclass[letterpaper, 10 pt, conference]{ieeeconf}
\IEEEoverridecommandlockouts
\usepackage{cite}
\usepackage{amsmath,amssymb,amsfonts}
\usepackage{algorithmic}
\usepackage{graphicx}
\usepackage{textcomp}
\usepackage{xcolor}
\usepackage{tikz}
\usetikzlibrary{patterns}
\usepackage{array}
\usepackage{hyperref}       
\usepackage{url}            
\usepackage{booktabs}       
\usepackage{amsfonts}       
\usepackage{nicefrac}       
\usepackage{microtype}      
\usepackage{xcolor}         
\let\labelindent\relax
\usepackage{enumitem}
\usepackage{graphicx}
\usepackage{caption}
\usepackage{subcaption}
\usepackage{transparent}
\usepackage{tikz}
\begin{document}

\newcommand{\customdata}{Matterport-Chair}
\newcommand{\ourmethod}{Sparse PointPillars}
\newcommand{\sparseone}{Sparse1+Dense23}
\newcommand{\sparsetwo}{Sparse12+Dense3}
\newcommand{\sparsewide}{Sparse+WideConv}

\newcommand{\jetsonhigh}{Xavier~High}
\newcommand{\jetsonlow}{Xavier~Low}

\newcommand{\desktop}{Desktop}
\newcommand{\robot}{Robot}

\newcommand{\desktopdense}{\desktop{} Dense}
\newcommand{\desktopsparse}{\desktop{} Sparse}
\newcommand{\jetsonhighsparse}{\jetsonhigh{} Sparse}
\newcommand{\jetsonhighdense}{\jetsonhigh{} Dense}
\newcommand{\jetsonlowsparse}{\jetsonlow{} Sparse}
\newcommand{\jetsonlowdense}{\jetsonlow{} Dense}
\newcommand{\robotdense}{\robot{} Dense}
\newcommand{\robotsparse}{\robot{} Sparse}

\newcommand{\valsize}{\scriptsize{}}
\newcommand{\pmsize}{\scriptsize{}}

\newcommand{\figref}[1]{Fig.~\ref{fig:#1}}
\newcommand{\figrefs}[2]{Figs.~\ref{fig:#1}--\ref{fig:#2}}
\newcommand{\figlabel}[1]{\label{fig:#1}} 
\newcommand{\tableref}[1]{Table~\ref{table:#1}}
\newcommand{\tablelabel}[1]{\label{table:#1}}
\newcommand{\sectionref}[1]{Section~\ref{section:#1}}
\newcommand{\sectionlabel}[1]{\label{section:#1}}
\newcommand{\appendixref}[1]{Appendix~\ref{appendix:#1}}
\newcommand{\appendixlabel}[1]{\label{appendix:#1}}

\newcommand{\tblheadsty}[1]{#1}
\newcommand{\densitywidth}{1.30in}

\title{\LARGE \bf \ourmethod{}: Maintaining and Exploiting Input Sparsity to Improve Runtime on Embedded Systems}

\author{Kyle Vedder$^{*}$ and Eric Eaton$^{\dagger}$
\thanks{$^{*}$Kyle Vedder is with the Dept.\ of Computer and Information Science, University of Pennsylvania, Philadelphia, PA 19104, USA
        {\tt\small kvedder@seas.upenn.edu}}%
\thanks{$^{\dagger}$Eric Eaton is with the Dept.\ of Computer and Information Science, University of Pennsylvania, Philadelphia, PA 19104, USA
        {\tt\small eeaton@seas.upenn.edu}}%
}


\usetikzlibrary{quotes}
\usetikzlibrary{decorations.pathreplacing}

\makeatletter
\newcommand{\thickhline}{%
    \noalign {\ifnum 0=`}\fi \hrule height 1pt
    \futurelet \reserved@a \@xhline
}
\newcolumntype{"}{@{\hskip\tabcolsep\vrule width 1pt\hskip\tabcolsep}}
\makeatother

\newcommand{\comEE}[1]{{\color{blue}EE: #1}}
\newcommand{\todo}[1]{{\color{red}TODO: #1}}
\newcommand{\tablespacehack}{\vspace{-0.0cm}}


\maketitle
\thispagestyle{empty}
\pagestyle{empty}

\begin{abstract}
Bird's Eye View (BEV) is a popular representation for processing 3D point clouds, and by its nature is  fundamentally sparse. Motivated by the computational limitations of mobile robot platforms, we create a fast, high-performance BEV 3D object detector that maintains \emph{and} exploits this input sparsity to decrease runtimes over non-sparse baselines and avoids the tradeoff between pseudoimage area and runtime. We present results on KITTI, a canonical 3D detection dataset, and \customdata{}, a novel Matterport3D-derived chair detection dataset from scenes in real furnished homes. We evaluate runtime characteristics using a desktop GPU, an embedded ML accelerator, and a robot CPU, demonstrating that our method results in significant detection speedups (2$\times$ or more) for embedded systems with only a modest decrease in detection quality. Our work represents a new approach for practitioners to optimize models for embedded systems by maintaining \emph{and} exploiting input sparsity throughout their entire pipeline to reduce runtime and resource usage while preserving detection performance. All models, weights, experimental configurations, and datasets used are publicly available\footnote{\url{https://vedder.io/sparse_point_pillars}}.
\end{abstract}

\section{Introduction}\sectionlabel{intro}

In modern robot perception systems, fitting state-of-the-art machine learning models within the power and compute constraints of on-board computational platforms is a major challenge. In the autonomous vehicle space, high-end desktop GPUs and CPUs are often available on-board, but this hardware still faces power and cost limits and must be shared with other components of the control stack. 
This challenge is even more pronounced for intelligent mobile robots;
for example, even a larger, high-end robot like the Fetch Freight~\cite{fetchfreight} is not able to power several desktop-grade GPUs and high-end CPUs in order to run its control stack. Instead, roboticists are forced to settle for smaller embedded systems like NVidia's Jetson~\cite{xavierwebinar}, while smaller platforms such as quadcopters often struggle to handle the weight or power requirements of even these embedded systems. 
This motivates the problem of developing machine learning models that have significantly reduced resource usage compared to existing models while preserving their performance --- models need to be shrunk not just to fit on smaller devices, but to fit while \emph{sharing} these resources with other components. 



In this paper we address this problem of reducing resource usage while preserving performance within PointPillars~\cite{pointpillars}, a point cloud-based 3D object detector that is \emph{very} popular among autonomous vehicle makers due to its speed and performance. PointPillars segregates the raw points into pillars and vectorizes these point collections into $N$ dimensional vectors,
resulting in a sparse Bird's Eye View (BEV) pseudoimage of the scene. PointPillars then processes the pseudoimage with a dense 2D convolutional Backbone and predicts bounding boxes using Single-Stage Detector~\cite{ssd}.
The PointPillars Backbone is the most expensive component of the pipeline, yet the pseudoimage it processes is highly sparse. As a result, large, empty sections of the image are unnecessarily convolved by the Backbone and runtimes directly scale with the pseudoimage area, presenting a trade-off.
%

\begin{figure*}[t!]
    \centering
\input{fig_pointpillarsarch_new}
\caption{Overview of PointPillars'~\cite{pointpillars} pipeline and \ourmethod{}' pipeline. Both use a shared pipeline which consumes a raw point cloud, pillarizes the scene (empty pillars are depicted in gray), \emph{gather}s the non-empty pillars into a sparse COO matrix, and vectorizes the pillars. PointPillars' Pillar Feature Net then \emph{scatter}s this sparse matrix into a dense matrix, passing this to its dense Backbone for processing. \ourmethod{}' Pillar Feature Net preserves the sparse COO representation, passing this to our sparse Backbone. Both PointPillars and \ourmethod{} use Single Stage Detector~\cite{ssd} to regress output bounding boxes. PointPillars spends the majority of its runtime running its dense Backbone; \ourmethod{} reduces this runtime by maintaining and exploiting the natural sparsity of the COO matrix via its new Pillar Feature Net and Backbone.}
    \figlabel{pointpillarsarch}
    \vspace{-1em}
\end{figure*}
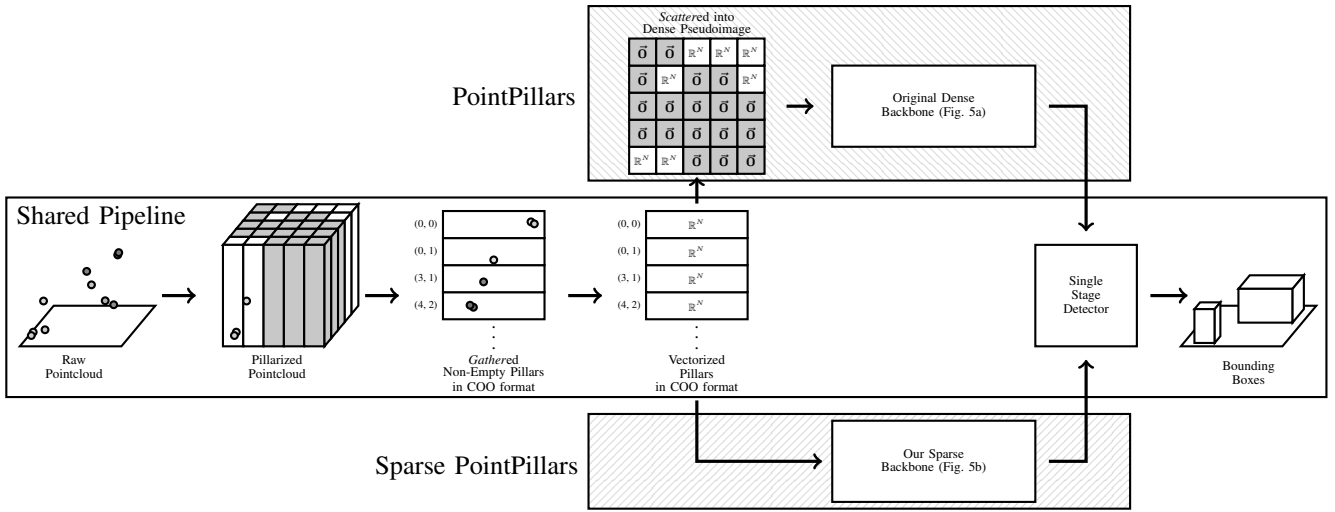

Our method, \ourmethod{}, avoids these inefficiencies and tradeoffs with a  novel processing pipeline that maintains \emph{and} exploits end-to-end sparsity to reduce runtime and resource usage on embedded systems while maintaining reasonable performance. Our main contributions include:

\begin{enumerate}[leftmargin=1.5em]
    \item A \textbf{new pipeline} that \textbf{maintains and exploits representational sparsity} and \textbf{avoids the trade-off between pseudoimage area and runtime}.
    \item \textbf{2$\times$ or more speedup on embedded systems}, allowing practitioners to gain large runtime speedups for the \textbf{same  power budget} or \textbf{modest runtime speedups for a significantly smaller power budget}, all in exchange for a modest decrease in detection quality.
    \item A general \textbf{design approach centered around representational sparsity} for efficient embedded system pipelines.
\end{enumerate}

An overview of the original PointPillars' pipeline and our new pipeline (\ourmethod{}) is shown in \figref{pointpillarsarch}.





\section{Related Work}\sectionlabel{relatedwork}

Solutions to the problem of reducing resource usage of machine learning models in order to deploy on embedded hardware typically fall into three categories. One general-purpose solution to this problem, model quantization~\cite{mixedprecision,quantizationgeneric1,quantizationgeneric2,distquantization,binaryquantization}, first trains models in a standard fashion using floating point weights and then, after training, converts some~\cite{mixedprecision} or all~\cite{quantizationgeneric1,quantizationgeneric2} weights into integer~\cite{distquantization} or binary~\cite{binaryquantization} quantized values that are faster to multiply than floating point values.  The quantized network is then finetuned, resulting in similar performance while running faster on accelerators (e.g. GPUs~\cite{tfquantization}), low-end compute devices (e.g. mobile phones~\cite{Wu_2016_CVPR}), or specialized hardware (e.g. FPGAs~\cite{pointpillarsquant}). 

A second model-agnostic approach, often paired with quantization, is model \emph{weight} sparsification, also known as model pruning. As per the Lottery Ticket Hypothesis~\cite{lotteryticket}, most weights contribute little to performance~\cite{optimalbraindamage,optimalbrainsurgeon,oldpruningsurvey} and can be pruned to improve runtime~\cite{stateofpruning,amperesparsity,structuredsparsity1,channelpruning,unstructuredsparsity1,unstructredsparsity2,unstructuredsparsity3}, either with regular structure to exploit hardware properties~\cite{structuredsparsity1,amperesparsity,channelpruning}, or without structure~\cite{unstructuredsparsity1,unstructredsparsity2,unstructuredsparsity3}, without much performance loss.

A third approach is to exploit data representations to reduce the computational burden, such as input representational sparsity. For example, a common approach for 3D object detection in point clouds is to voxelize the 3D space and perform 3D convolutions through a pipeline similar to 2D object detection~\cite{voxelnet,second,rangedilated}. 3D convolution of these voxels dominates network runtime~\cite{voxelnet}, but the voxels tend to be highly sparse; point clouds from KITTI contain over 100,000 points, but when voxelized into 16cm cubes, over 95\% are empty~\cite{second,pointpillars}. This motivates sparsity-aware 3D convolutional methods such as SECOND~\cite{second}, which performs sparse 3D convolutions to produce mathematically identical results with significantly less computation and runtime. SBNet~\cite{sbnet}, a BEV object detector, takes a spiritually similar approach by doing dense 2D convolutions only in coarsely masked regions of relevance to reduce the area convolved, decreasing runtime without a significant performance hit. 

Our method, \ourmethod{}, falls into this third category, using sparse matrix representations and multiple types of specialized 2D sparse convolutions to decrease runtime without a significant impact on performance. Importantly, our input sparsity-aware contributions are \emph{orthogonal} to model quantization and weight sparsification, thus providing practitioners another useful tool in improving model runtime.

\section{Maintaining and Exploiting End-to-End Sparsity within \ourmethod{}}\sectionlabel{pipelinemodifications}

This section 
describes \ourmethod{}'s new processing pipeline. We then theoretically analyze (\sectionref{theory}) and empirically validate (\sectionref{evaluation}) our modifications.

\subsection{New Pillar Feature Net}\sectionlabel{featurenet}

PointPillars performs pillerization (\figref{pointpillarsarch}) via its Pillar Feature Net which \emph{gather}s the non-zero entries of the full scene into a coordinate (COO) format sparse tensor~\cite{cootensors} with a fixed number of pillars and a fixed number of points per pillar (set a priori), along with the coordinate location of each pillar. The Feature Net then vectorizes each pillar using a PointNet-like vectorizer~\cite{pointnet}. 
In the original PointPillars' Feature Net, these resulting vectors are then \emph{scatter}ed back into a standard dense tensor in the shape of the full scene.

\begin{figure*}[t!]
\centering
\begin{minipage}[t]{0.99\columnwidth}
  \input{fig_pseudoimage_baseline}
\end{minipage}%
\hfill
\begin{minipage}[t]{0.99\columnwidth}
  \input{fig_pseudoimage_sparse}
\end{minipage}
\end{figure*}

In our modified Pillar Feature Net, we elide this \emph{scatter} step, thus maintaining the sparse COO matrix format of the vectorized pillars. This significantly reduces GPU requirements by avoiding an additional allocation of a \mbox{$\mathrm{scene\_width} \times \mathrm{scene\_height} \times N$} matrix plus complex matrix masking to insert the appropriate values, \emph{and} allows the Pillar Feature Net to emit a sparse pseudoimage output, enabling its further exploitation by our new Backbone.

\subsection{New Backbone}\sectionlabel{backbone}

The original PointPillars Backbone (\figref{backbonesbaseline}) takes in the dense tensor format pseudoimage from the \emph{scatter} step and processes it with a convolutional feature pyramid network~\cite{featurepyramidnetworks} style Backbone. This Backbone emits a single large pseudoimage of half the width and height of the input pseudoimage, composed of intermediary pseudoimages from the three layers of the Backbone concatenated along the channel axis.
Due to the heavy use of standard $3 \times 3$ stride-1 convolutions throughout the Backbone, there is significant smearing of non-zero entries across the pseudoimage; \figref{baselinepseudoimages} visualizes this for an input from the KITTI dataset: \figref{baselineinput} shows the input pseudoimage, and \figrefs{baselineconv1}{baselineconv3} show the smearing of non-zero entries across zero entries in the pseudoimage as it travels through the Backbone. 

Due to this smearing, na\"ively modifying the original PointPillars Backbone to perform sparse convolutions~\cite{second} on the sparse COO format pseudoimage from our new Pillar Feature Net would fail to \emph{preserve} input sparsity. A sparse convolution (\figref{convcompareblur}) is mathematically equal to a standard convolution (zero entries are simply skipped to save computation), and thus later pseudoimages would require convolution of almost all entries, e.g.\ \figref{baselineconv3}, hurting runtime.

Thus, \ourmethod{}' new Backbone (\figref{backbonesours}) operates on the sparse COO format pseudoimage with a combination of carefully placed sparse convolutions and submanifold (SubM) convolutions~\cite{submanifoldconv}, a type of convolution that only convolves filters centered on existing non-zero entries in order to maintain \emph{and} exploit the pseudoimage's sparsity throughout the Backbone.  \figref{convcompare} shows the results of these different types of convolutions, and \figref{sparsepseudoimages} visually demonstrates their impact on the pseudoimage's sparsity as it flows through the \ourmethod{} Backbone. Compared to PointPillars' Backbone, \ourmethod{}' Backbone maintains and exploits sparsity in three key ways: 1) its \texttt{BatchNorm} is only applied to the recorded non-zero entries, leaving the zero entries unaffected (this is trivial for COO tensors as only non-zero values are recorded), 2) it replaces the $3 \times 3$ stride-2 convolutions with $2\times 2$ stride-2 convolutions when shrinking the pseudoimage, ensuring that non-zero entries in the higher resolution pseudoimage appear only once in the lower resolution pseudoimage, and 3)~it~replaces the $3 \times 3$ stride-1 convolutions with $3 \times 3$ stride-1 SubM convolutions. 
As we will show, these changes require the new Backbone to perform fewer computations (\sectionref{theory}) resulting in significantly improved runtime in practice (\sectionref{evaluation}).


\input{fig_conv_compare}

\input{fig_backbones}

\section{Theoretical Analysis}
\sectionlabel{theory}

Our new Backbone maintains and exploits input sparsity to perform fewer operations and avoid the pseudoimage area vs runtime tradeoff in order to achieve faster runtime compared to PointPillars' Backbone.
For PointPillars, the number of convolutions performed by its Backbone (\figref{backbonesbaseline}) is a function of the area and number of channels of the input pseudoimage; the \emph{values} of the input pseudoimage are irrelevant. By comparison, the number of convolutions performed by \ourmethod{}' Backbone (\figref{backbonesours}) is a function of the area, number of channels, \emph{and} pseudoimage density (the fraction of non-zero values of the input pseudoimage). Due to the strategic use of sparse $2 \times 2$ and SubM stride-$2$ convolutions in each \texttt{Conv} block, \ourmethod{}' Backbone increases the pseudoimage density by \emph{at most} $4 \times $ per block; however, in practice this density increase is far below $4 \times $ as non-zero entries tend to cluster near one another (e.g.\ \figref{sparsepseudoimages}). 

\tableref{opstable} outlines  the type and number of convolutions performed by each Backbone, providing an exact count for PointPillars and an \emph{upper bound} for \ourmethod{}. When the input pseudoimage density $D$ approaches $1$, \ourmethod{}' Backbone converges to PointPillars' Backbone with the $3\times 3$ stride-$2$ convolutions replaced with $2 \times 2$ stride-$2$ convolutions; when $D$ approaches $0$, \ourmethod{}' Backbone performs significantly fewer convolutions. 

In practice, input density $D$ is very small for realistic data; for KITTI's test data, the median $D$ is $0.02459$ (min $0.013321$, max $0.03899$) and for \customdata{}'s test data, the median $D$ is $0.00750$ (min $0.00029$, max $0.01679$). Translating this to convolution operations count, for KITTI's median density, our Backbone performs \emph{at least} 50\% fewer convolutions, and for \customdata{}'s median density, our Backbone performs \emph{at least} 79\% fewer convolutions.

As \ourmethod{}' runtime is dependent on input density $D$ and PointPillars' is not, \ourmethod{} avoids the tradeoff between pseudoimage area and runtime, providing practitioners more flexibility in tuning their detector without significant runtime increases. For example, practitioners want to  decrease the pillar size to allow for the capture of finer-grained objects 
or increase in the maximum sensor range to detect of objects further away (both of which decrease $D$ as much as increase $W\times H$), and \ourmethod{} enables both of these without the same runtime hit as PointPillars.



\begin{table}[h!]
\caption{Number of convolutions performed by PointPillars' Backbones and an \emph{upper bound} on number of convolutions performed by  \ourmethod{}' Backbone, for an input pseudoimage of size $W \times H$ with $C$ channels and $D$ density.}
\begin{tabular}{l|l|l}
\tblheadsty{Operation}             & \tblheadsty{Baseline Count} & \tblheadsty{\ourmethod{}'s Upper Bound}\\ \hline   
\rule{0pt}{14pt}$3 \times 3$ Conv              & $\frac{15}{4}C^2HW$     & \begin{tabular}[c]{@{}l@{}}$C^2 H W (\min(\frac{3}{4}, 3 D)$ \\ $+ \min(\frac{5}{4}, 20 D) +  \min(\frac{5}{4}, 80 D))$\end{tabular} \\ 
\rule{0pt}{14pt}$2 \times 2$ Conv              & $0$                     & \begin{tabular}[c]{@{}l@{}}$C^2 H W (\frac{D}{4} +  \min(\frac{1}{8}, \frac{1}{2} D)$\\ $+ \min(\frac{1}{8}, 2D))$\end{tabular}      \\
\rule{0pt}{10pt}$1 \times 1$ Conv$^\mathsf{T}$ & $\frac{1}{2}C^2HW$      & $C^2HW \min(\frac{1}{2}, 2D)$                                                                                                       \\
\rule{0pt}{10pt}$2 \times 2$ Conv$^\mathsf{T}$ & $\frac{1}{4}C^2HW$      & $C^2HW \min(\frac{1}{4}, 4D)$                                                                                                      \\
\rule{0pt}{10pt}$4 \times 4$ Conv$^\mathsf{T}$ & $\frac{1}{8}C^2HW$      & $C^2HW \min(\frac{1}{8}, 8D)$                                                                                                     
\end{tabular}
\tablelabel{opstable}
\end{table}

\section{Empirical Evaluation}\sectionlabel{evaluation}

To validate the design of \ourmethod{}, we implemented it in \texttt{Open3D-ML}~\cite{open3d}, a high-quality third party implementation of PointPillars, using the Minkowski Engine~\cite{minkowskiengine} for sparse convolutions. To demonstrate \ourmethod{}' value on embedded systems via a realistic task, we compare it against PointPillars on \emph{\customdata{}}, a custom chair detection task derived from Matterport3D~\cite{matterport}, an indoor 3D scan dataset designed to simulate a task required of real service robots (\sectionref{matterport}). We evaluate the two trained \customdata{} models across three compute platforms: a desktop with a high-end GPU, an embedded ML accelerator configured for minimal and maximal power modes, and the CPU of a high-end commercial robot.

To contextualize \ourmethod{}' performance, we evaluate it against PointPillars on the KITTI~\cite{kittiobj} dataset (\sectionref{kitti}), a well understood 3D self-driving dataset used for evaluation in the original PointPillars paper, using a high-end GPU desktop setup common in these evaluations, and compare these results to other baseline models. We also perform several ablative studies on our Backbone to demonstrate that it produces a reasonable trade-off between runtime and detection performance.

\input{table_customdataset_runtimes}

\input{table_kitti_perf}

\input{table_kitti_runtimes}

\subsection{\customdata{} Evaluation with Embedded Performance}\sectionlabel{matterport}

To simulate a realistic detection task faced by a service robot or other embodied platform using embedded compute systems, we constructed a labeled chair detection dataset \emph{\customdata{}} using point clouds and their object labels sampled from houses in Matterport3D~\cite{matterport}. Matterport3D is a dataset of multiple building-scale indoor 3D meshes constructed using many high-resolution panoramic RGBD views taken inside real houses and labeled with 3D bounding boxes and semantic labels for over 20 different object classes. To generate our training and test dataset, we sampled point clouds of random views from the perspective of a robot sitting one meter off the ground across four different high quality house meshes, producing a train/test split of 7,500 point clouds each (the same size as the KITTI splits). We post-processed the bounding boxes, aligning them vertically and rejecting boxes that were highly occluded, associated with too few points, or caused by dataset noise such as holes in the house mesh. The training and test splits along with the generation code are available on the project webpage.

Both PointPillars and \ourmethod{} are trained with 5cm$\times$5cm pillars and $768 \times 512$ pillar pseudoimage (set using the max absolute \emph{X} and \emph{Y}-axis values of the training data point clouds), $66\%$/$33\%$ train/validation splits, and standard hyperparameters, with the exception that \ourmethod{} performs 50 more epochs in order to converge. Despite performing 25\% more epochs, \ourmethod{} trains 1.8$\times$ faster. Our evaluation follows the KITTI protocol of measuring the average precision (AP) at a detection threshold of 50\% Intersection over Union (IoU) of the bounding box relative to ground truth on two key benchmarks: the bounding boxes from BEV (\emph{BEV AP}) and in full 3D (\emph{3D AP}).

\ourmethod{} lags behind PointPillars in performance by 6.04\% AP on \emph{BEV} and by 4.61\% AP on \emph{3D} (PointPillars achieved 84.09\% AP on \emph{BEV} and 80.66\% AP on \emph{3D}). However, as shown in \tableref{customtable}, due to \customdata{}'s low density, \ourmethod{} is significantly faster than PointPillars across the full range of compute platforms available to embodied agents: a desktop with an AMD Ryzen 7 3700X CPU and an NVidia 2080ti GPU (denoted \emph{\desktop{}}), an NVidia Jetson Xavier embedded ML accelerator configured for the highest and lowest power settings (30 Watt, 8 core mode denoted \emph{\jetsonhigh{}} and 10 Watt, 2 core mode denoted \emph{\jetsonlow{}}), and a Fetch Freight~\cite{fetchfreight} robot's built-in four core Intel i5-4590S CPU (denoted \emph{\robot{}}). 
\ourmethod{} is more than $1.5\times$ as fast as PointPillars on \emph{\desktop{}} (fast enough for 60Hz inference), more than $2\times$ as fast on \emph{\jetsonhigh{}} (fast enough for 10Hz inference), almost $3\times$ as fast on \emph{\jetsonlow{}} (fast enough for 6Hz inference), and more than $4\times$ as fast on \emph{\robot{}} (fast enough for 4Hz inference). 

Of note, the recorded Backbone runtimes for \ourmethod{} on GPU accelerated platforms is \emph{slower} than PointPillars, but the BBox Extract stage is \emph{faster} despite using \emph{identical} code. The runtime difference comes from the time taken to allocate the memory for the anchor boxes---both allocate the same size GPU array, but due to pipelining and earlier memory cleanup that inflated the \ourmethod{}' Backbone's runtime, it is able to allocate the final anchor boxes faster. The \emph{\robot{}} evaluations demonstrate that when GPU pipelining is not a factor, \ourmethod{}' Backbone is far faster than PointPillars' Backbone, and the BBox Extract stage runs at roughly the same speed.

\subsection{KITTI Evaluation with Ablative Studies}\sectionlabel{kitti}

KITTI~\cite{kittiobj}, a self-driving car dataset of LiDAR point clouds with human-annotated 3D bounding boxes, is a common benchmark dataset in 3D object detection and is used as the evaluation dataset in the original  PointPillars paper. Both \ourmethod{} and PointPillars are trained on the KITTI Car detection task, configured with the default 16cm$\times$16cm pillars, $504 \times 440$ pillar pseudoimage, 50\%/50\% train/validation split, and hyperparameter configurations outlined in the PointPillars paper, with the exception that \ourmethod{} performs 50 more epochs in order to converge. Despite performing 25\% more epochs, \ourmethod{} trains in roughly the same amount of time. Our evaluation follows the prescribed KITTI evaluation protocol of measuring the average precision (AP) at a detection threshold of 70\% Intersection over Union (IoU) of the bounding box relative to ground truth on two key benchmarks: the bounding boxes from a BEV (\emph{BEV AP}) and the full 3D bounding boxes (\emph{3D AP}). KITTI does not have public labels for its test set, so in keeping with the literature~\cite{voxelnet,second,pointpillars} we report results on the validation set. Results are separated for the three KITTI difficulty levels (Easy, Medium, Hard), and runtimes are recorded on a dedicated desktop with an AMD Ryzen 7 3700X CPU and an NVidia 2080ti GPU.

Additionally, to better understand our contributions, we perform two ablative studies to answer these questions:

\begin{enumerate}
\item Would making the new Backbone only partially sparse provide a better performance-runtime trade-off? 
\item Would approximating the flow of smeared information with wider convolutions improve performance?
\end{enumerate} 

To answer 1), we replace the later sections of our new Backbone with their dense counterparts from the original Backbone to construct two variants. Using \figref{backbones}'s \texttt{Conv} block definitions, the ablated variant \emph{\sparseone} uses the sparse \texttt{Conv} block 1 and dense \texttt{Conv} blocks 2 and 3, and the variant \emph{\sparsetwo} uses sparse \texttt{Conv} blocks 1 and 2 with a dense \texttt{Conv} block 3.  To answer 2), we modify the filter size of the first SubM convolution of each \texttt{Conv} block to be $9\times9$ in order to simulate the information transfer caused by pseudoimage smearing in the original Backbone. We refer to this variant as  \emph{\sparsewide}.

The absolute percentage of Average Precision (\% AP) 
for PointPillars on each benchmark and the relative performance of \ourmethod{} and its ablations are shown in \tableref{perfbaselinesparse}. Relative to PointPillars, \ourmethod{} performs roughly 5\% AP worse on \emph{BEV} and roughly 8.5\% AP worse on \emph{3D}, and roughly equally to the ablative models, with \sparseone{} performing slightly better and with \sparsetwo{} and \sparsewide{} performing worse. Together, these results indicate that SubM convolutional blocks in the Backbone are more difficult to train, even if the block has access to the same information as the dense model.

The runtime for each component of each method is reported in \tableref{runtimes}, with our \ourmethod{} running 0.18ms faster than PointPillars. Our Feature Net runs 0.18ms faster as it avoids the \emph{scatter} step, but like with \customdata{}, the recorded runtime for our Backbone is actually 5.22ms \emph{slower} than PointPillars Backbone and the BBox Extract stage with its \emph{identical} code is 5.22ms \emph{faster} due to memory pipelining. Unsurprisingly, \sparseone{} and \sparsetwo{} are both slower than PointPillars and \ourmethod{} due to the Backbone pipelining interruption when converting from a sparse to a dense tensor, and \sparsewide{} is significantly slower due to its very large convolutions.

To contextualize \ourmethod{}' 14.2ms runtime vs PointPillars' 14.4ms runtime in \tableref{perfbaselinesparse}, we refer to the reported KITTI runtime numbers for other sparsity based 3D detection approaches discussed in \sectionref{relatedwork}. SECOND~\cite{second}, a sparse 3D convolution-based object detector, reports a runtime of 50ms for its large variant (which the authors use to evaluate performance) and 25ms for its small variant using $20cm\times20cm\times40cm$ voxels. SBNet~\cite{sbnet}, a 2D BEV object detector that performs dense convolution inside coarse pseudolabel masks, reports 17.9ms runtime using $10cm\times10cm$ pillars. Direct head-to-head performance and runtime comparisons against prior art can be found in the PointPillars~\cite{pointpillars} paper.


\section{Conclusion and Future Work}\sectionlabel{futurework}

This work demonstrates that \ourmethod{} 
allows practitioners to trade small amounts of model performance for significant decreases in runtime and resource usage on embedded systems. For example, on our \customdata{} dataset, \textbf{\ourmethod{} runs faster on the Jetson Xavier in low power mode than PointPillars does in high power mode}, allowing a practitioner to save power \emph{and} get reduced runtimes at the cost of a few \% AP. Alternatively, PointPillars runs at less than 1Hz on the robot's CPU; \textbf{with \ourmethod{}, practitioners can reliably run at 1Hz \emph{and} have more than 75\% of the CPU budget left to run other components of the robot control stack}. By providing faster runtimes via our architectural design, \ourmethod{} provides practitioners new tools in their toolbox to build and optimize their full control stack.

This work can be extend by exploring model quantization and weight pruning in tandem with our new pipeline. Prior art has shown significant quantization of PointPillars results in only minor drops in performance~\cite{pointpillarsquant}. When combined with \ourmethod{}, this may result in significant further reductions in runtime for a modest drop in performance, or enable inference on more exotic hardware, e.g. FPGAs.

Additionally, this work would benefit from further performance evaluation using a Streaming AP~\cite{streamingap} style measure extended to 3D detectors. In this work, we evaluated detection quality with AP, a standard metric in the vision literature that matches output detections to the \emph{input} point cloud. However, this evaluation protocol does not represent the problem practitioners face:  in dynamic environments, the detection is most useful if it matches the state of the world \emph{at the time it is emitted}. The world changes while the detector is performing inference and so a quick, lower quality detection is potentially better representative of the world upon emission than a slow, higher quality detection. A streaming measure would directly consider the dramatic latency reductions of \ourmethod{} in the evaluation of its accuracy, better reflecting the problem formulation that practitioners face.

Finally, sparse perception pipelines, such as this work, are (and will continue to be) \textbf{directly impactful for autonomous vehicle makers}. New LiDAR sensors offer increased detection ranges ~\cite{nuscenesdataset,Argoverse,argoversetwo}, causing increased pseudoimage area and reduced density, making \ourmethod{}' 
sparsity exploitation   even more important in maintaining low runtimes.

\section*{Acknowledgments}
We gratefully acknowledge the useful feedback on this work from Jorge Mendez and Marcel Hussing. 
This research was partially supported by the DARPA Lifelong Learning Machines program under grant FA8750-18-2-0117, the DARPA SAIL-ON program under contract HR001120C0040, and the Army Research Office under MURI grant W911NF20-1-0080. 

\bibliographystyle{./bibliography/IEEEtran}
\bibliography{mybib}

\end{document}

%% file: fig_pointpillarsarch_new.tex
\begin{tikzpicture}[thick, scale=0.9]
    \path[draw, color=black, fill=white]
        (-0.2, -0.75)
        -- (19.3, -0.75)
        -- (19.3, 2.2)
        -- (-0.2, 2.2)
        -- cycle;
    \path[draw, color=black]    (9.55, 0.7250000000000001)    node[] () {{$\textup{}$}};
    \path[draw, color=black]    (1.2, 1.9)    node[] () {{$\textup{Shared Pipeline}$}};
    \path[draw, color=black, fill=white, pattern=north west lines, pattern color={rgb,255:red,220;green,220;blue,220}]
        (8.4, 2.45)
        -- (16.4, 2.45)
        -- (16.4, 5.03)
        -- (8.4, 5.03)
        -- cycle;
    \path[draw, color=black]    (12.399999999999999, 3.74)    node[] () {{$\textup{}$}};
    \path[draw, color=black]    (7.300000000000001, 3.7)    node[] () {{$\textup{PointPillars}$}};
    \path[draw, color=black, fill=white, pattern=north east lines, pattern color={rgb,255:red,220;green,220;blue,220}]
        (8.4, -2.4)
        -- (16.4, -2.4)
        -- (16.4, -1.0)
        -- (8.4, -1.0)
        -- cycle;
    \path[draw, color=black]    (12.399999999999999, -1.7)    node[] () {{$\textup{}$}};
    \path[draw, color=black]    (6.750000000000001, -1.8)    node[] () {{$\textup{Sparse PointPillars}$}};
    \path[draw, color=black, fill=white]
        (0.0, 0.0)
        -- (1.5, 0.0)
        -- (2.0, 0.6)
        -- (0.5, 0.6)
        -- cycle;
    \path[draw, color=black]    (1.0, 0.3)    node[] () {{$\textup{}$}};
    \path[draw, fill={rgb,255:red,223.2000849849862;green,223.2000849849862;blue,223.2000849849862}]    (0.19694663362445602, 0.20978201000275148)    circle (0.05);
    \path[draw, fill={rgb,255:red,214.01604511443594;green,214.01604511443594;blue,214.01604511443594}]    (0.3495368767122503, 0.6698369135890041)    circle (0.05);
    \path[draw, fill={rgb,255:red,205.5425623033479;green,205.5425623033479;blue,205.5425623033479}]    (0.1690813265137962, 0.15949845816103608)    circle (0.05);
    \path[draw, fill={rgb,255:red,201.23129035468583;green,201.23129035468583;blue,201.23129035468583}]    (0.37177574133645935, 0.2447948880731765)    circle (0.05);
    \path[draw, fill={rgb,255:red,196.91889518493315;green,196.91889518493315;blue,196.91889518493315}]    (1.0579297940134453, 0.9089746496202993)    circle (0.05);
    \path[draw, fill={rgb,255:red,175.98813395746848;green,175.98813395746848;blue,175.98813395746848}]    (1.2607053457051718, 0.6737043211450198)    circle (0.05);
    \path[draw, fill={rgb,255:red,141.61176654634005;green,141.61176654634005;blue,141.61176654634005}]    (1.3860799333938891, 0.6147487034516415)    circle (0.05);
    \path[draw, fill={rgb,255:red,131.91662805607749;green,131.91662805607749;blue,131.91662805607749}]    (0.98748657380068, 1.108080070326596)    circle (0.05);
    \path[draw, fill={rgb,255:red,123.40643716419302;green,123.40643716419302;blue,123.40643716419302}]    (1.4457791928631574, 1.3512253773484397)    circle (0.05);
    \path[draw, fill={rgb,255:red,118.68622744408404;green,118.68622744408404;blue,118.68622744408404}]    (1.4607741194991952, 1.3854054276352556)    circle (0.05);
    \path[draw, color=black]    (0.8, -0.20000000000000004)    node[font=\fontsize{5pt}{5pt}] () {{\fontsize{5pt}{5pt}$\textup{Raw}$}}    (0.8, -0.4)    node[font=\fontsize{5pt}{5pt}] () {{\fontsize{5pt}{5pt}$\textup{Pointcloud}$}};
    \path[->, very thick, draw]    (2.1, 0.75)    -- (2.6, 0.75);
    \path[draw, color=black, fill=white]
        (3.0, 1.5)
        -- (3.3, 1.5)
        -- (3.4, 1.62)
        -- (3.1, 1.62)
        -- cycle;
    \path[draw, color=black]    (3.1999999999999997, 1.56)    node[] () {{$\textup{}$}};
    \path[draw, color=black, fill={rgb,255:red,200;green,200;blue,200}]
        (3.1, 1.62)
        -- (3.4, 1.62)
        -- (3.5, 1.7400000000000002)
        -- (3.2, 1.7400000000000002)
        -- cycle;
    \path[draw, color=black]    (3.3, 1.6800000000000002)    node[] () {{$\textup{}$}};
    \path[draw, color=black, fill={rgb,255:red,200;green,200;blue,200}]
        (3.2, 1.74)
        -- (3.5, 1.74)
        -- (3.6, 1.8599999999999999)
        -- (3.3000000000000003, 1.8599999999999999)
        -- cycle;
    \path[draw, color=black]    (3.4000000000000004, 1.7999999999999998)    node[] () {{$\textup{}$}};
    \path[draw, color=black, fill={rgb,255:red,200;green,200;blue,200}]
        (3.3, 1.8599999999999999)
        -- (3.5999999999999996, 1.8599999999999999)
        -- (3.6999999999999997, 1.98)
        -- (3.4, 1.98)
        -- cycle;
    \path[draw, color=black]    (3.5, 1.92)    node[] () {{$\textup{}$}};
    \path[draw, color=black, fill={rgb,255:red,200;green,200;blue,200}]
        (3.4, 1.98)
        -- (3.6999999999999997, 1.98)
        -- (3.8, 2.1)
        -- (3.5, 2.1)
        -- cycle;
    \path[draw, color=black]    (3.5999999999999996, 2.04)    node[] () {{$\textup{}$}};
    \path[draw, color=black, fill=white]
        (3.3, 1.5)
        -- (3.5999999999999996, 1.5)
        -- (3.6999999999999997, 1.62)
        -- (3.4, 1.62)
        -- cycle;
    \path[draw, color=black]    (3.5, 1.56)    node[] () {{$\textup{}$}};
    \path[draw, color=black, fill={rgb,255:red,200;green,200;blue,200}]
        (3.4, 1.62)
        -- (3.6999999999999997, 1.62)
        -- (3.8, 1.7400000000000002)
        -- (3.5, 1.7400000000000002)
        -- cycle;
    \path[draw, color=black]    (3.5999999999999996, 1.6800000000000002)    node[] () {{$\textup{}$}};
    \path[draw, color=black, fill={rgb,255:red,200;green,200;blue,200}]
        (3.5, 1.74)
        -- (3.8, 1.74)
        -- (3.9, 1.8599999999999999)
        -- (3.6, 1.8599999999999999)
        -- cycle;
    \path[draw, color=black]    (3.6999999999999997, 1.7999999999999998)    node[] () {{$\textup{}$}};
    \path[draw, color=black, fill=white]
        (3.6, 1.8599999999999999)
        -- (3.9, 1.8599999999999999)
        -- (4.0, 1.98)
        -- (3.7, 1.98)
        -- cycle;
    \path[draw, color=black]    (3.8, 1.92)    node[] () {{$\textup{}$}};
    \path[draw, color=black, fill={rgb,255:red,200;green,200;blue,200}]
        (3.7, 1.98)
        -- (4.0, 1.98)
        -- (4.1, 2.1)
        -- (3.8000000000000003, 2.1)
        -- cycle;
    \path[draw, color=black]    (3.9000000000000004, 2.04)    node[] () {{$\textup{}$}};
    \path[draw, color=black, fill={rgb,255:red,200;green,200;blue,200}]
        (3.6, 1.5)
        -- (3.9, 1.5)
        -- (4.0, 1.62)
        -- (3.7, 1.62)
        -- cycle;
    \path[draw, color=black]    (3.8, 1.56)    node[] () {{$\textup{}$}};
    \path[draw, color=black, fill={rgb,255:red,200;green,200;blue,200}]
        (3.7, 1.62)
        -- (4.0, 1.62)
        -- (4.1, 1.7400000000000002)
        -- (3.8000000000000003, 1.7400000000000002)
        -- cycle;
    \path[draw, color=black]    (3.9000000000000004, 1.6800000000000002)    node[] () {{$\textup{}$}};
    \path[draw, color=black, fill={rgb,255:red,200;green,200;blue,200}]
        (3.8, 1.74)
        -- (4.1, 1.74)
        -- (4.199999999999999, 1.8599999999999999)
        -- (3.9, 1.8599999999999999)
        -- cycle;
    \path[draw, color=black]    (3.9999999999999996, 1.7999999999999998)    node[] () {{$\textup{}$}};
    \path[draw, color=black, fill={rgb,255:red,200;green,200;blue,200}]
        (3.9, 1.8599999999999999)
        -- (4.2, 1.8599999999999999)
        -- (4.3, 1.98)
        -- (4.0, 1.98)
        -- cycle;
    \path[draw, color=black]    (4.1, 1.92)    node[] () {{$\textup{}$}};
    \path[draw, color=black, fill=white]
        (4.0, 1.98)
        -- (4.3, 1.98)
        -- (4.3999999999999995, 2.1)
        -- (4.1, 2.1)
        -- cycle;
    \path[draw, color=black]    (4.199999999999999, 2.04)    node[] () {{$\textup{}$}};
    \path[draw, color=black, fill={rgb,255:red,200;green,200;blue,200}]
        (3.9, 1.5)
        -- (4.2, 1.5)
        -- (4.3, 1.62)
        -- (4.0, 1.62)
        -- cycle;
    \path[draw, color=black]    (4.1, 1.56)    node[] () {{$\textup{}$}};
    \path[draw, color=black, fill={rgb,255:red,200;green,200;blue,200}]
        (4.0, 1.62)
        -- (4.3, 1.62)
        -- (4.3999999999999995, 1.7400000000000002)
        -- (4.1, 1.7400000000000002)
        -- cycle;
    \path[draw, color=black]    (4.199999999999999, 1.6800000000000002)    node[] () {{$\textup{}$}};
    \path[draw, color=black, fill={rgb,255:red,200;green,200;blue,200}]
        (4.1, 1.74)
        -- (4.3999999999999995, 1.74)
        -- (4.499999999999999, 1.8599999999999999)
        -- (4.199999999999999, 1.8599999999999999)
        -- cycle;
    \path[draw, color=black]    (4.3, 1.7999999999999998)    node[] () {{$\textup{}$}};
    \path[draw, color=black, fill={rgb,255:red,200;green,200;blue,200}]
        (4.2, 1.8599999999999999)
        -- (4.5, 1.8599999999999999)
        -- (4.6, 1.98)
        -- (4.3, 1.98)
        -- cycle;
    \path[draw, color=black]    (4.3999999999999995, 1.92)    node[] () {{$\textup{}$}};
    \path[draw, color=black, fill=white]
        (4.3, 1.98)
        -- (4.6, 1.98)
        -- (4.699999999999999, 2.1)
        -- (4.3999999999999995, 2.1)
        -- cycle;
    \path[draw, color=black]    (4.499999999999999, 2.04)    node[] () {{$\textup{}$}};
    \path[draw, color=black, fill={rgb,255:red,200;green,200;blue,200}]
        (4.2, 1.5)
        -- (4.5, 1.5)
        -- (4.6, 1.62)
        -- (4.3, 1.62)
        -- cycle;
    \path[draw, color=black]    (4.3999999999999995, 1.56)    node[] () {{$\textup{}$}};
    \path[draw, color=black, fill={rgb,255:red,200;green,200;blue,200}]
        (4.3, 1.62)
        -- (4.6, 1.62)
        -- (4.699999999999999, 1.7400000000000002)
        -- (4.3999999999999995, 1.7400000000000002)
        -- cycle;
    \path[draw, color=black]    (4.499999999999999, 1.6800000000000002)    node[] () {{$\textup{}$}};
    \path[draw, color=black, fill={rgb,255:red,200;green,200;blue,200}]
        (4.4, 1.74)
        -- (4.7, 1.74)
        -- (4.8, 1.8599999999999999)
        -- (4.5, 1.8599999999999999)
        -- cycle;
    \path[draw, color=black]    (4.6000000000000005, 1.7999999999999998)    node[] () {{$\textup{}$}};
    \path[draw, color=black, fill=white]
        (4.5, 1.8599999999999999)
        -- (4.8, 1.8599999999999999)
        -- (4.8999999999999995, 1.98)
        -- (4.6, 1.98)
        -- cycle;
    \path[draw, color=black]    (4.699999999999999, 1.92)    node[] () {{$\textup{}$}};
    \path[draw, color=black, fill=white]
        (4.6, 1.98)
        -- (4.8999999999999995, 1.98)
        -- (4.999999999999999, 2.1)
        -- (4.699999999999999, 2.1)
        -- cycle;
    \path[draw, color=black]    (4.8, 2.04)    node[] () {{$\textup{}$}};
    \path[draw, color=black, fill=white]
        (3.0, 1.5)
        -- (3.3, 1.5)
        -- (3.3, 0.0)
        -- (3.0, 0.0)
        -- cycle;
    \path[draw, color=black]    (3.15, 0.75)    node[] () {{$\textup{}$}};
    \path[draw, color=black, fill=white]
        (3.3, 1.5)
        -- (3.5999999999999996, 1.5)
        -- (3.5999999999999996, 0.0)
        -- (3.3, 0.0)
        -- cycle;
    \path[draw, color=black]    (3.45, 0.75)    node[] () {{$\textup{}$}};
    \path[draw, color=black, fill={rgb,255:red,200;green,200;blue,200}]
        (3.6, 1.5)
        -- (3.9, 1.5)
        -- (3.9, 0.0)
        -- (3.6, 0.0)
        -- cycle;
    \path[draw, color=black]    (3.75, 0.75)    node[] () {{$\textup{}$}};
    \path[draw, color=black, fill={rgb,255:red,200;green,200;blue,200}]
        (3.9, 1.5)
        -- (4.2, 1.5)
        -- (4.2, 0.0)
        -- (3.9, 0.0)
        -- cycle;
    \path[draw, color=black]    (4.05, 0.75)    node[] () {{$\textup{}$}};
    \path[draw, color=black, fill={rgb,255:red,200;green,200;blue,200}]
        (4.2, 1.5)
        -- (4.5, 1.5)
        -- (4.5, 0.0)
        -- (4.2, 0.0)
        -- cycle;
    \path[draw, color=black]    (4.35, 0.75)    node[] () {{$\textup{}$}};
    \path[draw, color=black, fill={rgb,255:red,200;green,200;blue,200}]
        (4.5, 1.5)
        -- (4.5, 0.0)
        -- (4.6, 0.12)
        -- (4.6, 1.62)
        -- cycle;
    \path[draw, color=black]    (4.55, 0.81)    node[] () {{$\textup{}$}};
    \path[draw, color=black, fill={rgb,255:red,200;green,200;blue,200}]
        (4.6, 1.62)
        -- (4.6, 0.1200000000000001)
        -- (4.699999999999999, 0.2400000000000001)
        -- (4.699999999999999, 1.7400000000000002)
        -- cycle;
    \path[draw, color=black]    (4.6499999999999995, 0.9300000000000002)    node[] () {{$\textup{}$}};
    \path[draw, color=black, fill={rgb,255:red,200;green,200;blue,200}]
        (4.7, 1.74)
        -- (4.7, 0.24)
        -- (4.8, 0.36)
        -- (4.8, 1.8599999999999999)
        -- cycle;
    \path[draw, color=black]    (4.75, 1.0499999999999998)    node[] () {{$\textup{}$}};
    \path[draw, color=black, fill=white]
        (4.8, 1.8599999999999999)
        -- (4.8, 0.3599999999999999)
        -- (4.8999999999999995, 0.47999999999999987)
        -- (4.8999999999999995, 1.98)
        -- cycle;
    \path[draw, color=black]    (4.85, 1.17)    node[] () {{$\textup{}$}};
    \path[draw, color=black, fill=white]
        (4.9, 1.98)
        -- (4.9, 0.48)
        -- (5.0, 0.6)
        -- (5.0, 2.1)
        -- cycle;
    \path[draw, color=black]    (4.95, 1.29)    node[] () {{$\textup{}$}};
    \path[draw, fill={rgb,255:red,223.2000849849862;green,223.2000849849862;blue,223.2000849849862}]    (3.196946633624456, 0.20978201000275148)    circle (0.05);
    \path[draw, fill={rgb,255:red,214.01604511443594;green,214.01604511443594;blue,214.01604511443594}]    (3.3495368767122504, 0.6698369135890041)    circle (0.05);
    \path[draw, fill={rgb,255:red,205.5425623033479;green,205.5425623033479;blue,205.5425623033479}]    (3.1690813265137963, 0.15949845816103608)    circle (0.05);
    \path[draw, color=black]    (3.8, -0.20000000000000004)    node[font=\fontsize{5pt}{5pt}] () {{\fontsize{5pt}{5pt}$\textup{Pillarized}$}}    (3.8, -0.4)    node[font=\fontsize{5pt}{5pt}] () {{\fontsize{5pt}{5pt}$\textup{Pointcloud}$}};
    \path[->, very thick, draw]    (5.1, 0.75)    -- (5.6, 0.75);
    \path[draw, color=black, fill=white]
        (6.25, 0.4)
        -- (7.75, 0.4)
        -- (7.75, 0.8)
        -- (6.25, 0.8)
        -- cycle;
    \path[draw, color=black]    (7.0, 0.6000000000000001)    node[] () {{$\textup{}$}};
    \path[draw, color=black, fill=white]
        (6.25, 0.8)
        -- (7.75, 0.8)
        -- (7.75, 1.2000000000000002)
        -- (6.25, 1.2000000000000002)
        -- cycle;
    \path[draw, color=black]    (7.0, 1.0)    node[] () {{$\textup{}$}};
    \path[draw, color=black, fill=white]
        (6.25, 1.2000000000000002)
        -- (7.75, 1.2000000000000002)
        -- (7.75, 1.6)
        -- (6.25, 1.6)
        -- cycle;
    \path[draw, color=black]    (7.0, 1.4)    node[] () {{$\textup{}$}};
    \path[draw, color=black, fill=white]
        (6.25, 1.6)
        -- (7.75, 1.6)
        -- (7.75, 2.0)
        -- (6.25, 2.0)
        -- cycle;
    \path[draw, color=black]    (7.0, 1.8)    node[] () {{$\textup{}$}};
    \path[draw, color=black]    (6.0, 1.8000000000000003)    node[font=\fontsize{4pt}{4pt}] () {{\fontsize{4pt}{4pt}$\textup{(0, 0)}$}};
    \path[draw, color=black]    (6.0, 1.4000000000000004)    node[font=\fontsize{4pt}{4pt}] () {{\fontsize{4pt}{4pt}$\textup{(0, 1)}$}};
    \path[draw, color=black]    (6.0, 1.0000000000000002)    node[font=\fontsize{4pt}{4pt}] () {{\fontsize{4pt}{4pt}$\textup{(3, 1)}$}};
    \path[draw, color=black]    (6.0, 0.6000000000000001)    node[font=\fontsize{4pt}{4pt}] () {{\fontsize{4pt}{4pt}$\textup{(4, 2)}$}};
    \path[draw, fill={rgb,255:red,115.0;green,115.0;blue,115.0}]    (6.7, 0.5775000000000003)    circle (0.05);
    \path[draw, fill={rgb,255:red,115.0;green,115.0;blue,115.0}]    (6.65, 0.6075000000000004)    circle (0.05);
    \path[draw, fill={rgb,255:red,135.0;green,135.0;blue,135.0}]    (6.85, 0.9525000000000002)    circle (0.05);
    \path[draw, fill={rgb,255:red,225.0;green,225.0;blue,225.0}]    (7.0, 1.2775000000000003)    circle (0.05);
    \path[draw, fill={rgb,255:red,225.0;green,225.0;blue,225.0}]    (7.55, 1.8400000000000003)    circle (0.05);
    \path[draw, fill={rgb,255:red,205.0;green,205.0;blue,205.0}]    (7.6, 1.8100000000000003)    circle (0.05);
    \path[draw, color=black]    (7.0, 0.25)    node[] () {{$\textup{\scriptsize\vdots}$}};
    \path[draw, color=black]
        (7.0, -0.2)    node[font=\fontsize{5pt}{5pt}] () {{\fontsize{5pt}{5pt}$\textup{\emph{Gather}ed}$}}
        (7.0, -0.4)    node[font=\fontsize{5pt}{5pt}] () {{\fontsize{5pt}{5pt}$\textup{Non-Empty Pillars}$}}
        (7.0, -0.6000000000000001)    node[font=\fontsize{5pt}{5pt}] () {{\fontsize{5pt}{5pt}$\textup{in COO format}$}};
    \path[->, very thick, draw]    (8.1, 0.75)    -- (8.6, 0.75);
    \path[draw, color=black, fill=white]
        (9.25, 0.4)
        -- (10.75, 0.4)
        -- (10.75, 0.8)
        -- (9.25, 0.8)
        -- cycle;
    \path[draw, color=black]    (10.0, 0.6000000000000001)    node[] () {{$\textup{}$}};
    \path[draw, color=black, fill=white]
        (9.25, 0.8)
        -- (10.75, 0.8)
        -- (10.75, 1.2000000000000002)
        -- (9.25, 1.2000000000000002)
        -- cycle;
    \path[draw, color=black]    (10.0, 1.0)    node[] () {{$\textup{}$}};
    \path[draw, color=black, fill=white]
        (9.25, 1.2000000000000002)
        -- (10.75, 1.2000000000000002)
        -- (10.75, 1.6)
        -- (9.25, 1.6)
        -- cycle;
    \path[draw, color=black]    (10.0, 1.4)    node[] () {{$\textup{}$}};
    \path[draw, color=black, fill=white]
        (9.25, 1.6)
        -- (10.75, 1.6)
        -- (10.75, 2.0)
        -- (9.25, 2.0)
        -- cycle;
    \path[draw, color=black]    (10.0, 1.8)    node[] () {{$\textup{}$}};
    \path[draw, color=black]    (9.0, 1.8000000000000003)    node[font=\fontsize{4pt}{4pt}] () {{\fontsize{4pt}{4pt}$\textup{(0, 0)}$}};
    \path[draw]    (10.0, 1.8000000000000003)    node[] () {{\fontsize{4pt}{4pt}$\mathbb{R}^{ \scalebox{0.5}{$ \scriptscriptstyle N$}}$}};
    \path[draw, color=black]    (9.0, 1.4000000000000004)    node[font=\fontsize{4pt}{4pt}] () {{\fontsize{4pt}{4pt}$\textup{(0, 1)}$}};
    \path[draw]    (10.0, 1.4000000000000004)    node[] () {{\fontsize{4pt}{4pt}$\mathbb{R}^{ \scalebox{0.5}{$ \scriptscriptstyle N$}}$}};
    \path[draw, color=black]    (9.0, 1.0000000000000002)    node[font=\fontsize{4pt}{4pt}] () {{\fontsize{4pt}{4pt}$\textup{(3, 1)}$}};
    \path[draw]    (10.0, 1.0000000000000002)    node[] () {{\fontsize{4pt}{4pt}$\mathbb{R}^{ \scalebox{0.5}{$ \scriptscriptstyle N$}}$}};
    \path[draw, color=black]    (9.0, 0.6000000000000001)    node[font=\fontsize{4pt}{4pt}] () {{\fontsize{4pt}{4pt}$\textup{(4, 2)}$}};
    \path[draw]    (10.0, 0.6000000000000001)    node[] () {{\fontsize{4pt}{4pt}$\mathbb{R}^{ \scalebox{0.5}{$ \scriptscriptstyle N$}}$}};
    \path[draw, color=black]    (10.0, 0.25)    node[] () {{$\textup{\scriptsize\vdots}$}};
    \path[draw, color=black]
        (10.0, -0.2)    node[font=\fontsize{5pt}{5pt}] () {{\fontsize{5pt}{5pt}$\textup{Vectorized}$}}
        (10.0, -0.4)    node[font=\fontsize{5pt}{5pt}] () {{\fontsize{5pt}{5pt}$\textup{Pillars}$}}
        (10.0, -0.6000000000000001)    node[font=\fontsize{5pt}{5pt}] () {{\fontsize{5pt}{5pt}$\textup{in COO format}$}};
    \path[->, very thick, draw]    (10.0, 2.11)    -- (10.0, 2.5);
    \path[draw, color=black, fill=white]
        (9.0, 2.55)
        -- (9.4, 2.55)
        -- (9.4, 2.9499999999999997)
        -- (9.0, 2.9499999999999997)
        -- cycle;
    \path[draw, color=black]    (9.2, 2.7499999999999996)    node[font=\fontsize{4pt}{4pt}] () {{\fontsize{4pt}{4pt}$\textup{$\mathbb{R}^{\scalebox{0.5}{$ \scriptscriptstyle N$}}$}$}};
    \path[draw, color=black, fill={rgb,255:red,200;green,200;blue,200}]
        (9.0, 2.9499999999999997)
        -- (9.4, 2.9499999999999997)
        -- (9.4, 3.3499999999999996)
        -- (9.0, 3.3499999999999996)
        -- cycle;
    \path[draw, color=black]    (9.2, 3.15)    node[font=\fontsize{4pt}{4pt}] () {{\fontsize{4pt}{4pt}$\textup{$\vec{\mathbf{0}}$}$}};
    \path[draw, color=black, fill={rgb,255:red,200;green,200;blue,200}]
        (9.0, 3.3499999999999996)
        -- (9.4, 3.3499999999999996)
        -- (9.4, 3.7499999999999996)
        -- (9.0, 3.7499999999999996)
        -- cycle;
    \path[draw, color=black]    (9.2, 3.55)    node[font=\fontsize{4pt}{4pt}] () {{\fontsize{4pt}{4pt}$\textup{$\vec{\mathbf{0}}$}$}};
    \path[draw, color=black, fill={rgb,255:red,200;green,200;blue,200}]
        (9.0, 3.75)
        -- (9.4, 3.75)
        -- (9.4, 4.15)
        -- (9.0, 4.15)
        -- cycle;
    \path[draw, color=black]    (9.2, 3.95)    node[font=\fontsize{4pt}{4pt}] () {{\fontsize{4pt}{4pt}$\textup{$\vec{\mathbf{0}}$}$}};
    \path[draw, color=black, fill={rgb,255:red,200;green,200;blue,200}]
        (9.0, 4.15)
        -- (9.4, 4.15)
        -- (9.4, 4.550000000000001)
        -- (9.0, 4.550000000000001)
        -- cycle;
    \path[draw, color=black]    (9.2, 4.3500000000000005)    node[font=\fontsize{4pt}{4pt}] () {{\fontsize{4pt}{4pt}$\textup{$\vec{\mathbf{0}}$}$}};
    \path[draw, color=black, fill=white]
        (9.4, 2.55)
        -- (9.8, 2.55)
        -- (9.8, 2.9499999999999997)
        -- (9.4, 2.9499999999999997)
        -- cycle;
    \path[draw, color=black]    (9.600000000000001, 2.7499999999999996)    node[font=\fontsize{4pt}{4pt}] () {{\fontsize{4pt}{4pt}$\textup{$\mathbb{R}^{\scalebox{0.5}{$ \scriptscriptstyle N$}}$}$}};
    \path[draw, color=black, fill={rgb,255:red,200;green,200;blue,200}]
        (9.4, 2.9499999999999997)
        -- (9.8, 2.9499999999999997)
        -- (9.8, 3.3499999999999996)
        -- (9.4, 3.3499999999999996)
        -- cycle;
    \path[draw, color=black]    (9.600000000000001, 3.15)    node[font=\fontsize{4pt}{4pt}] () {{\fontsize{4pt}{4pt}$\textup{$\vec{\mathbf{0}}$}$}};
    \path[draw, color=black, fill={rgb,255:red,200;green,200;blue,200}]
        (9.4, 3.3499999999999996)
        -- (9.8, 3.3499999999999996)
        -- (9.8, 3.7499999999999996)
        -- (9.4, 3.7499999999999996)
        -- cycle;
    \path[draw, color=black]    (9.600000000000001, 3.55)    node[font=\fontsize{4pt}{4pt}] () {{\fontsize{4pt}{4pt}$\textup{$\vec{\mathbf{0}}$}$}};
    \path[draw, color=black, fill=white]
        (9.4, 3.75)
        -- (9.8, 3.75)
        -- (9.8, 4.15)
        -- (9.4, 4.15)
        -- cycle;
    \path[draw, color=black]    (9.600000000000001, 3.95)    node[font=\fontsize{4pt}{4pt}] () {{\fontsize{4pt}{4pt}$\textup{$\mathbb{R}^{\scalebox{0.5}{$ \scriptscriptstyle N$}}$}$}};
    \path[draw, color=black, fill={rgb,255:red,200;green,200;blue,200}]
        (9.4, 4.15)
        -- (9.8, 4.15)
        -- (9.8, 4.550000000000001)
        -- (9.4, 4.550000000000001)
        -- cycle;
    \path[draw, color=black]    (9.600000000000001, 4.3500000000000005)    node[font=\fontsize{4pt}{4pt}] () {{\fontsize{4pt}{4pt}$\textup{$\vec{\mathbf{0}}$}$}};
    \path[draw, color=black, fill={rgb,255:red,200;green,200;blue,200}]
        (9.8, 2.55)
        -- (10.200000000000001, 2.55)
        -- (10.200000000000001, 2.9499999999999997)
        -- (9.8, 2.9499999999999997)
        -- cycle;
    \path[draw, color=black]    (10.0, 2.7499999999999996)    node[font=\fontsize{4pt}{4pt}] () {{\fontsize{4pt}{4pt}$\textup{$\vec{\mathbf{0}}$}$}};
    \path[draw, color=black, fill={rgb,255:red,200;green,200;blue,200}]
        (9.8, 2.9499999999999997)
        -- (10.200000000000001, 2.9499999999999997)
        -- (10.200000000000001, 3.3499999999999996)
        -- (9.8, 3.3499999999999996)
        -- cycle;
    \path[draw, color=black]    (10.0, 3.15)    node[font=\fontsize{4pt}{4pt}] () {{\fontsize{4pt}{4pt}$\textup{$\vec{\mathbf{0}}$}$}};
    \path[draw, color=black, fill={rgb,255:red,200;green,200;blue,200}]
        (9.8, 3.3499999999999996)
        -- (10.200000000000001, 3.3499999999999996)
        -- (10.200000000000001, 3.7499999999999996)
        -- (9.8, 3.7499999999999996)
        -- cycle;
    \path[draw, color=black]    (10.0, 3.55)    node[font=\fontsize{4pt}{4pt}] () {{\fontsize{4pt}{4pt}$\textup{$\vec{\mathbf{0}}$}$}};
    \path[draw, color=black, fill={rgb,255:red,200;green,200;blue,200}]
        (9.8, 3.75)
        -- (10.200000000000001, 3.75)
        -- (10.200000000000001, 4.15)
        -- (9.8, 4.15)
        -- cycle;
    \path[draw, color=black]    (10.0, 3.95)    node[font=\fontsize{4pt}{4pt}] () {{\fontsize{4pt}{4pt}$\textup{$\vec{\mathbf{0}}$}$}};
    \path[draw, color=black, fill=white]
        (9.8, 4.15)
        -- (10.200000000000001, 4.15)
        -- (10.200000000000001, 4.550000000000001)
        -- (9.8, 4.550000000000001)
        -- cycle;
    \path[draw, color=black]    (10.0, 4.3500000000000005)    node[font=\fontsize{4pt}{4pt}] () {{\fontsize{4pt}{4pt}$\textup{$\mathbb{R}^{\scalebox{0.5}{$ \scriptscriptstyle N$}}$}$}};
    \path[draw, color=black, fill={rgb,255:red,200;green,200;blue,200}]
        (10.2, 2.55)
        -- (10.6, 2.55)
        -- (10.6, 2.9499999999999997)
        -- (10.2, 2.9499999999999997)
        -- cycle;
    \path[draw, color=black]    (10.399999999999999, 2.7499999999999996)    node[font=\fontsize{4pt}{4pt}] () {{\fontsize{4pt}{4pt}$\textup{$\vec{\mathbf{0}}$}$}};
    \path[draw, color=black, fill={rgb,255:red,200;green,200;blue,200}]
        (10.2, 2.9499999999999997)
        -- (10.6, 2.9499999999999997)
        -- (10.6, 3.3499999999999996)
        -- (10.2, 3.3499999999999996)
        -- cycle;
    \path[draw, color=black]    (10.399999999999999, 3.15)    node[font=\fontsize{4pt}{4pt}] () {{\fontsize{4pt}{4pt}$\textup{$\vec{\mathbf{0}}$}$}};
    \path[draw, color=black, fill={rgb,255:red,200;green,200;blue,200}]
        (10.2, 3.3499999999999996)
        -- (10.6, 3.3499999999999996)
        -- (10.6, 3.7499999999999996)
        -- (10.2, 3.7499999999999996)
        -- cycle;
    \path[draw, color=black]    (10.399999999999999, 3.55)    node[font=\fontsize{4pt}{4pt}] () {{\fontsize{4pt}{4pt}$\textup{$\vec{\mathbf{0}}$}$}};
    \path[draw, color=black, fill={rgb,255:red,200;green,200;blue,200}]
        (10.2, 3.75)
        -- (10.6, 3.75)
        -- (10.6, 4.15)
        -- (10.2, 4.15)
        -- cycle;
    \path[draw, color=black]    (10.399999999999999, 3.95)    node[font=\fontsize{4pt}{4pt}] () {{\fontsize{4pt}{4pt}$\textup{$\vec{\mathbf{0}}$}$}};
    \path[draw, color=black, fill=white]
        (10.2, 4.15)
        -- (10.6, 4.15)
        -- (10.6, 4.550000000000001)
        -- (10.2, 4.550000000000001)
        -- cycle;
    \path[draw, color=black]    (10.399999999999999, 4.3500000000000005)    node[font=\fontsize{4pt}{4pt}] () {{\fontsize{4pt}{4pt}$\textup{$\mathbb{R}^{\scalebox{0.5}{$ \scriptscriptstyle N$}}$}$}};
    \path[draw, color=black, fill={rgb,255:red,200;green,200;blue,200}]
        (10.6, 2.55)
        -- (11.0, 2.55)
        -- (11.0, 2.9499999999999997)
        -- (10.6, 2.9499999999999997)
        -- cycle;
    \path[draw, color=black]    (10.8, 2.7499999999999996)    node[font=\fontsize{4pt}{4pt}] () {{\fontsize{4pt}{4pt}$\textup{$\vec{\mathbf{0}}$}$}};
    \path[draw, color=black, fill={rgb,255:red,200;green,200;blue,200}]
        (10.6, 2.9499999999999997)
        -- (11.0, 2.9499999999999997)
        -- (11.0, 3.3499999999999996)
        -- (10.6, 3.3499999999999996)
        -- cycle;
    \path[draw, color=black]    (10.8, 3.15)    node[font=\fontsize{4pt}{4pt}] () {{\fontsize{4pt}{4pt}$\textup{$\vec{\mathbf{0}}$}$}};
    \path[draw, color=black, fill={rgb,255:red,200;green,200;blue,200}]
        (10.6, 3.3499999999999996)
        -- (11.0, 3.3499999999999996)
        -- (11.0, 3.7499999999999996)
        -- (10.6, 3.7499999999999996)
        -- cycle;
    \path[draw, color=black]    (10.8, 3.55)    node[font=\fontsize{4pt}{4pt}] () {{\fontsize{4pt}{4pt}$\textup{$\vec{\mathbf{0}}$}$}};
    \path[draw, color=black, fill=white]
        (10.6, 3.75)
        -- (11.0, 3.75)
        -- (11.0, 4.15)
        -- (10.6, 4.15)
        -- cycle;
    \path[draw, color=black]    (10.8, 3.95)    node[font=\fontsize{4pt}{4pt}] () {{\fontsize{4pt}{4pt}$\textup{$\mathbb{R}^{\scalebox{0.5}{$ \scriptscriptstyle N$}}$}$}};
    \path[draw, color=black, fill=white]
        (10.6, 4.15)
        -- (11.0, 4.15)
        -- (11.0, 4.550000000000001)
        -- (10.6, 4.550000000000001)
        -- cycle;
    \path[draw, color=black]    (10.8, 4.3500000000000005)    node[font=\fontsize{4pt}{4pt}] () {{\fontsize{4pt}{4pt}$\textup{$\mathbb{R}^{\scalebox{0.5}{$ \scriptscriptstyle N$}}$}$}};
    \path[draw, color=black]    (10.0, 4.88)    node[font=\fontsize{5pt}{5pt}] () {{\fontsize{5pt}{5pt}$\textup{\emph{Scatter}ed into}$}}    (10.0, 4.680000000000001)    node[font=\fontsize{5pt}{5pt}] () {{\fontsize{5pt}{5pt}$\textup{Dense Pseudoimage}$}};
    \path[->, very thick, draw]    (11.325, 3.55)    -- (11.725, 3.55);
    \path[draw, color=black, fill=white]
        (12.0, 2.9499999999999997)
        -- (15.0, 2.9499999999999997)
        -- (15.0, 4.1499999999999995)
        -- (12.0, 4.1499999999999995)
        -- cycle;
    \path[draw, color=black]    (13.5, 3.65)    node[font=\fontsize{5pt}{5pt}] () {{\fontsize{5pt}{5pt}$\textup{Original Dense}$}}    (13.5, 3.4499999999999997)    node[font=\fontsize{5pt}{5pt}] () {{\fontsize{5pt}{5pt}$\textup{Backbone (\figref{backbonesbaseline})}$}};
    \path[->, very thick, draw]
        (10.0, -0.8)
        -- (10.0, -1.7000000000000002)
        -- (11.85, -1.7000000000000002);
    \path[draw, color=black, fill=white]
        (12.0, -2.3000000000000003)
        -- (15.0, -2.3000000000000003)
        -- (15.0, -1.1000000000000003)
        -- (12.0, -1.1000000000000003)
        -- cycle;
    \path[draw, color=black]    (13.5, -1.6000000000000003)    node[font=\fontsize{5pt}{5pt}] () {{\fontsize{5pt}{5pt}$\textup{Our Sparse}$}}    (13.5, -1.8000000000000005)    node[font=\fontsize{5pt}{5pt}] () {{\fontsize{5pt}{5pt}$\textup{Backbone (\figref{backbonesours})}$}};
    \path[draw, color=black, fill=white]
        (15.0, 0.0)
        -- (16.5, 0.0)
        -- (16.5, 1.5)
        -- (15.0, 1.5)
        -- cycle;
    \path[draw, color=black]
        (15.75, 0.95)    node[font=\fontsize{5pt}{5pt}] () {{\fontsize{5pt}{5pt}$\textup{Single}$}}
        (15.75, 0.75)    node[font=\fontsize{5pt}{5pt}] () {{\fontsize{5pt}{5pt}$\textup{Stage}$}}
        (15.75, 0.55)    node[font=\fontsize{5pt}{5pt}] () {{\fontsize{5pt}{5pt}$\textup{Detector}$}};
    \path[->, very thick, draw]
        (15.2, 3.55)
        -- (15.75, 3.55)
        -- (15.75, 1.7);
    \path[->, very thick, draw]
        (15.2, -1.7000000000000002)
        -- (15.75, -1.7000000000000002)
        -- (15.75, -0.2);
    \path[->, very thick, draw]    (16.7, 0.75)    -- (17.25, 0.75);
    \path[draw, color=black, fill=white]
        (17.15, 0.0)
        -- (18.65, 0.0)
        -- (19.15, 0.6)
        -- (17.65, 0.6)
        -- cycle;
    \path[draw, color=black]    (18.15, 0.3)    node[] () {{$\textup{}$}};
    \path[draw, color=black, fill=white]
        (18.0, 0.85)
        -- (18.8, 0.85)
        -- (18.96666666666667, 1.05)
        -- (18.166666666666668, 1.05)
        -- cycle;
    \path[draw, color=black]    (18.483333333333334, 0.95)    node[] () {{$\textup{}$}};
    \path[draw, color=black, fill=white]
        (18.0, 0.85)
        -- (18.8, 0.85)
        -- (18.8, 0.35)
        -- (18.0, 0.35)
        -- cycle;
    \path[draw, color=black]    (18.4, 0.6)    node[] () {{$\textup{}$}};
    \path[draw, color=black, fill=white]
        (18.8, 0.85)
        -- (18.8, 0.35)
        -- (18.96666666666667, 0.55)
        -- (18.96666666666667, 1.05)
        -- cycle;
    \path[draw, color=black]    (18.883333333333333, 0.7)    node[] () {{$\textup{}$}};
    \path[draw, color=black, fill=white]
        (17.349999999999998, 0.55)
        -- (17.65, 0.55)
        -- (17.73333333333333, 0.65)
        -- (17.43333333333333, 0.65)
        -- cycle;
    \path[draw, color=black]    (17.541666666666664, 0.6)    node[] () {{$\textup{}$}};
    \path[draw, color=black, fill=white]
        (17.349999999999998, 0.55)
        -- (17.65, 0.55)
        -- (17.65, 0.050000000000000044)
        -- (17.349999999999998, 0.050000000000000044)
        -- cycle;
    \path[draw, color=black]    (17.5, 0.30000000000000004)    node[] () {{$\textup{}$}};
    \path[draw, color=black, fill=white]
        (17.65, 0.55)
        -- (17.65, 0.050000000000000044)
        -- (17.73333333333333, 0.15000000000000005)
        -- (17.73333333333333, 0.65)
        -- cycle;
    \path[draw, color=black]    (17.691666666666666, 0.35000000000000003)    node[] () {{$\textup{}$}};
    \path[draw, color=black]    (18.15, -0.30000000000000004)    node[font=\fontsize{5pt}{5pt}] () {{\fontsize{5pt}{5pt}$\textup{Bounding}$}}    (18.15, -0.5)    node[font=\fontsize{5pt}{5pt}] () {{\fontsize{5pt}{5pt}$\textup{Boxes}$}};
\end{tikzpicture}

%% file: fig_pseudoimage_baseline.tex
  \centering
  \begin{subfigure}[t]{0.45\columnwidth}
         \centering
         \frame{\includegraphics[height=\densitywidth, angle=90]{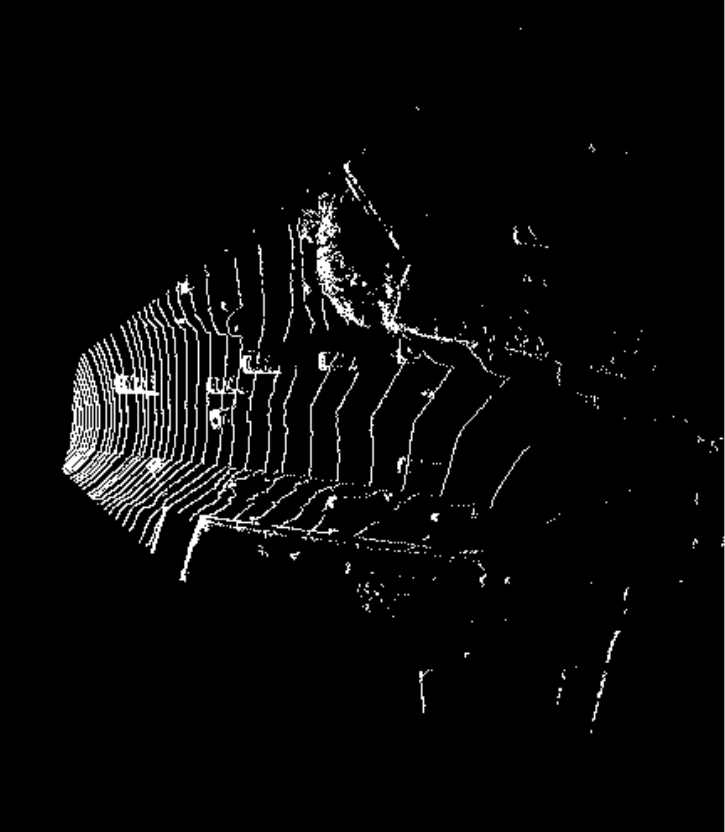}}
         \caption{Input pseudoimage}
         \figlabel{baselineinput}
     \end{subfigure}
     \hfill
     \begin{subfigure}[t]{0.45\columnwidth}
         \centering
         \frame{\includegraphics[height=\densitywidth, angle=90]{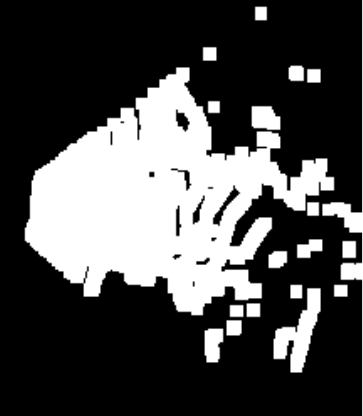}}
         \caption{After \texttt{Conv} block 1}
         \figlabel{baselineconv1}
     \end{subfigure}
     \\
     \begin{subfigure}[t]{0.45\columnwidth}
         \centering
         \frame{\includegraphics[height=\densitywidth, angle=90]{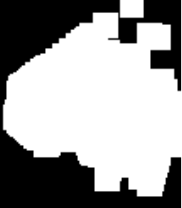}}
         \caption{After \texttt{Conv} block 2}
         \figlabel{baselineconv2}
     \end{subfigure}
     \hfill
     \begin{subfigure}[t]{0.45\columnwidth}
         \centering
         \frame{\includegraphics[height=\densitywidth, angle=90]{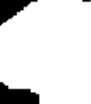}}
         \caption{After \texttt{Conv} block 3}
         \figlabel{baselineconv3}
     \end{subfigure}
  \caption{Pseudoimages from original PointPillars with \texttt{BatchNorm} removed for visualization; black represents zero entries on all channels and white represents at least one non-zero channel entry. With \texttt{BatchNorm} retained, sparsity is entirely destroyed as zero entries are modified.}
  \figlabel{baselinepseudoimages}

%% file: fig_pseudoimage_sparse.tex
  \centering
  \begin{subfigure}[t]{0.45\columnwidth}
         \centering
         \frame{\includegraphics[height=\densitywidth, angle=90]{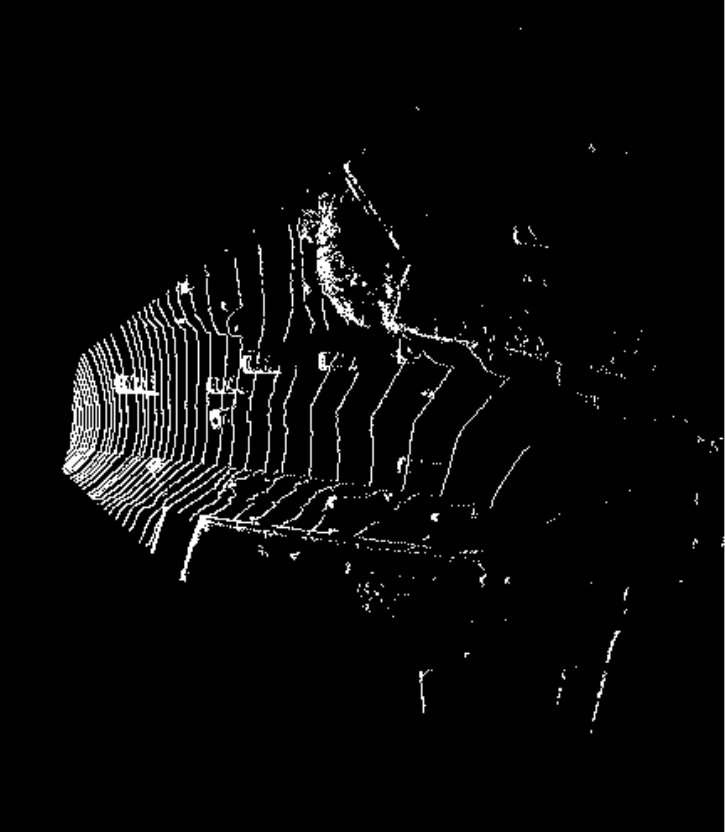}}
         \caption{Input pseudoimage}
     \end{subfigure}
     \hfill
     \begin{subfigure}[t]{0.45\columnwidth}
         \centering
         \frame{\includegraphics[height=\densitywidth, angle=90]{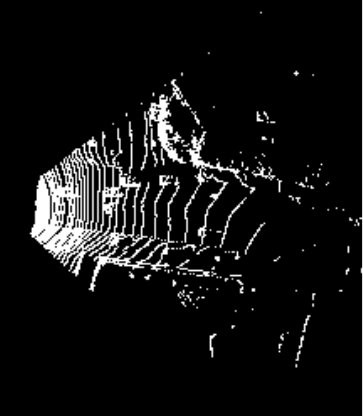}}
         \caption{After \texttt{Conv} block 1}
     \end{subfigure}
     \hfill
     \begin{subfigure}[t]{0.45\columnwidth}
         \centering
         \frame{\includegraphics[height=\densitywidth, angle=90]{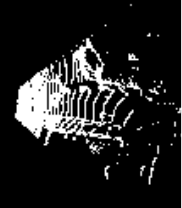}}
         \caption{After \texttt{Conv} block 2}
     \end{subfigure}
     \hfill
     \begin{subfigure}[t]{0.45\columnwidth}
         \centering
         \frame{\includegraphics[height=\densitywidth, angle=90]{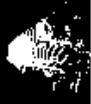}}
         \caption{After \texttt{Conv} block 3}
     \end{subfigure}
  \caption{Pseudoimages from our \ourmethod{} run on a sample from KITTI. Black represents zero entries on all channels and white represents at least one non-zero channel entry. Due to the use of SubM convs and \texttt{BatchNorm} only operating over non-zero entries, sparsity is maintained.} 
  \figlabel{sparsepseudoimages}

%% file: fig_conv_compare.tex
\newcommand{\convcomparewidth}{2cm}
\begin{figure}[htb!]
  \centering
  \begin{subfigure}[t]{0.32\columnwidth}
         \centering
         \includegraphics[width=\convcomparewidth]{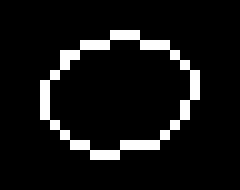}
         \caption{Input image}
         \figlabel{convcompareinput}
     \end{subfigure}%
     \hfill
     \begin{subfigure}[t]{0.32\columnwidth}
         \centering
         \includegraphics[width=\convcomparewidth]{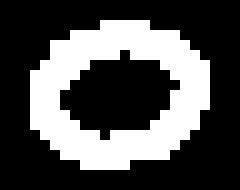}
         \captionsetup{justification=centering}
         \caption{After $3\times3$\\Standard Conv}
         \figlabel{convcompareblur}
     \end{subfigure}%
     \hfill
     \begin{subfigure}[t]{0.32\columnwidth}
         \centering
         \includegraphics[width=\convcomparewidth]{submanifold/img.png}
         \captionsetup{justification=centering}
         \caption{After $3\times3$\\SubM Conv}
         \figlabel{convcomparesubm}
     \end{subfigure}%
  \caption{$3\times3$ stride-1 Standard Convolution versus $3\times3$ Submanifold (SubM) Convolution. Black represents zero entries on all channels and white represents at least one non-zero channel entry. Standard convolutions can be centered on zero entries next to non-zero entries, resulting in a new non-zero entry, causing smearing and destroying spasity. SubM convolutions are only centered on non-zero entries, preventing smearing and maintaining sparsity.}
  \figlabel{convcompare}
\end{figure}

%% file: fig_backbones.tex
\newcommand{\backbonesscale}{0.75}
\newcommand{\imgheight}{0.34in}
\newcommand{\filterheight}{0.31in}
\newcommand{\filterwidth}{0.35in}
\newcommand{\filteropacity}{0.5}
\newcommand{\backbonestextreg}{\fontsize{6.0}{1}\linespread{0.2}\selectfont}
\newcommand{\backbonestexttiny}{\fontsize{4.5}{1}\selectfont}
\newcommand{\backbonestextrepeat}{\fontsize{7.1}{1}\selectfont}
\newcommand{\backbonestextblock}{\fontsize{7.1}{1}\selectfont}
\begin{figure}[htbp]
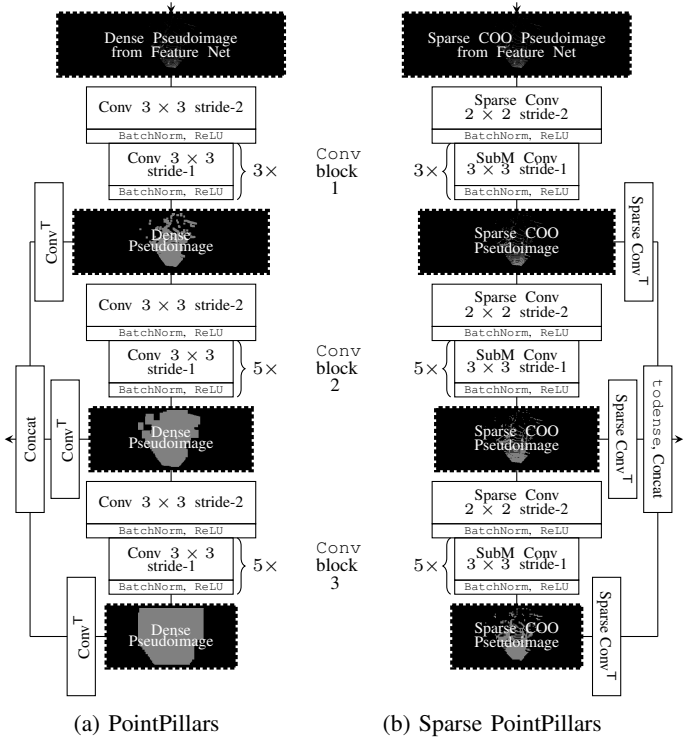

    \centering
    \hspace{-1.5em}
    \begin{subfigure}[b]{0.25\textwidth}
         \centering
         \begin{tikzpicture}[scale=\backbonesscale]
         \def\pimgsize{2.0}
         \draw [-stealth](0,-1.0) -- (0, -1.2);
        \draw[densely dotted,ultra thick,fill=black] (-\pimgsize,-1.2) rectangle (\pimgsize,-2.3);
        \path (-\pimgsize,-1.2) rectangle
        (\pimgsize,-2.3) node[pos=.5,align=center,text width=4.85cm] {\frame{\includegraphics[height=\imgheight,angle=90]{dense/before.pdf}}};
        \path (-\pimgsize,-1.2) rectangle
        (\pimgsize,-2.3) node[pos=.5,align=center,text width=4.85cm] {{
        \transparent{\filteropacity}\textcolor{black}{\rule{\filterwidth}{\filterheight}}
        }};;
        \path (-\pimgsize,-1.2) rectangle
        (\pimgsize,-2.3) node[pos=.5,align=center,text=white,text width=4.85cm] {\backbonestextreg Dense Pseudoimage\\from Feature Net\par};
    \foreach \idx in {1, 2, 3}{
        \def\stepdown{3.5}
        \def\y{\idx*-3.5}
        \def\numconv{\ifthenelse{\(\idx=1\)}{3}{5}}
        \ifthenelse{\(\idx=2\)}{\def\tconvlineend{-2.25}}{\def\tconvlineend{-2.5}};
        \draw[] (0,1+\y) -- (0,1.2+\y);
        \draw[] (-1.5,0.25+\y) rectangle (1.5,1+\y) node[pos=.5, text width=3*\backbonesscale cm,align=center] {\backbonestextreg Conv $3\times 3$ stride-2 \par};
        \draw[] (-1.5,0.25+\y) rectangle (1.5,0+\y) node[pos=.5, text width=3cm,align=center] {\backbonestexttiny \texttt{BatchNorm}, \texttt{ReLU}\par};
        \draw[] (-1.1,-0.75+\y) rectangle (1.1,0+\y) node[pos=.5, text width=2*\backbonesscale cm,align=center] {\backbonestextreg Conv $3\times 3$ stride-1 \par};
        \draw[] (-1.1,-0.75+\y) rectangle (1.1,-1+\y) node[pos=.5, text width=2*\backbonesscale cm,align=center] {\backbonestexttiny \texttt{BatchNorm}, \texttt{ReLU}\par};
        \draw [decorate,decoration={brace,mirror,amplitude=3pt},xshift=2pt,yshift=0pt]
    (1.1,-1+\y) -- (1.1,0+\y) node [black,midway,xshift=11] 
    {\backbonestextrepeat $\numconv{} \times$};
        \draw[] (0,-1+\y) -- (0,-1.2+\y);
        \draw[densely dotted,ultra thick, fill=black] (-\pimgsize+\idx/\stepdown,-1.2+\y) rectangle (\pimgsize-\idx/\stepdown,-2.3+\y);
        \path (-\pimgsize+\idx/\stepdown,-1.2+\y) rectangle (\pimgsize-\idx/\stepdown,-2.3+\y) node[pos=.5, text width=2*\backbonesscale cm,align=center] {\frame{\includegraphics[height=\imgheight,angle=90]{dense/block\idx.pdf}}};
        \path (-\pimgsize+\idx/\stepdown,-1.2+\y) rectangle (\pimgsize-\idx/\stepdown,-2.3+\y) node[pos=.5,align=center] {{
        \transparent{\filteropacity}\textcolor{black}{\rule{\filterwidth}{\filterheight}}
        }};
        \path (-\pimgsize+\idx/\stepdown,-1.2+\y) rectangle (\pimgsize-\idx/\stepdown,-2.3+\y) node[pos=.5, text=white, text width=2*\backbonesscale cm,align=center] {\backbonestextreg Dense Pseudoimage\par};
        \draw[] (-\pimgsize+\idx/\stepdown,-1.75+\y) -- (-\pimgsize+\idx/\stepdown-0.2,-1.75+\y);
        \def\convtxoff{-\pimgsize+\idx/\stepdown - 0.2};
        \draw[] (\convtxoff,-2.8+\y) rectangle (\convtxoff-0.5,-0.7+\y)
        node[pos=.5,align=center, rotate=90] {\backbonestextreg Conv$^\mathsf{T}$\par};
        \draw[] (\convtxoff-0.5,-1.75+\y) -- (\tconvlineend,-1.75+\y);
        \ifnum\idx=1
            \draw[] (\tconvlineend,-1.75+\y) -- (\tconvlineend,-7.45);
        \fi
        \ifnum\idx=2
            \draw[] (\tconvlineend,-3.05+\y) rectangle (\tconvlineend-0.5,-0.45+\y) node[pos=.5,align=center, rotate=90] {\backbonestextreg Concat\par};
            \draw [-stealth](\tconvlineend-0.5,-1.75+\y) -- (\tconvlineend-0.7,-1.75+\y);
        \fi
        \ifnum\idx=3
            \draw[] (\tconvlineend,-1.75+\y) -- (\tconvlineend,-10.05);
        \fi
    }
    \end{tikzpicture}
         \caption{PointPillars}
         \figlabel{backbonesbaseline}
     \end{subfigure}
     \hfill
     \begin{subfigure}[b]{0.25\textwidth}
         \centering
         \hspace{-1.5em}
         \begin{tikzpicture}[scale=\backbonesscale]
         \def\pimgsize{2.0}
            \draw [-stealth](0,-1.0) -- (0, -1.2);
            \draw[densely dotted,ultra thick, fill=black] (-\pimgsize,-1.2) rectangle (\pimgsize,-2.3);
            \path (-\pimgsize,-1.2) rectangle (\pimgsize,-2.3) node[pos=.5,align=center,text width=4.85cm] {\frame{\includegraphics[height=\imgheight,angle=90]{dense/before.pdf}}};
            \path (-\pimgsize,-1.2) rectangle (\pimgsize,-2.3) node[pos=.5,align=center,text width=4.85cm] {{
        \transparent{\filteropacity}\textcolor{black}{\rule{\filterwidth}{\filterheight}}
        }};
            \path (-\pimgsize,-1.2) rectangle (\pimgsize,-2.3) node[pos=.5,align=center,text width=4.85cm,text=white] {\backbonestextreg Sparse COO Pseudoimage\\from Feature Net\par};
            \foreach \idx in {1, 2, 3}{
                \def\stepdown{3.5};
                \def\y{\idx*-3.5};
                \def\numconv{\ifthenelse{\(\idx=1\)}{3}{5}};
                \ifthenelse{\(\idx=2\)}{\def\tconvlineend{2.25}}{\def\tconvlineend{2.5}};
                \draw[] (0,1+\y) -- (0,1.2+\y);
                \draw[] (-1.5,0.25+\y) rectangle (1.5,1+\y) node[pos=.5, text width=3*\backbonesscale cm,align=center] {\backbonestextreg Sparse Conv $2\times 2$ stride-2\par};
                \draw[] (-1.5,0.25+\y) rectangle (1.5,0+\y) node[pos=.5, text width=3*\backbonesscale cm,align=center] {\backbonestexttiny \texttt{BatchNorm}, \texttt{ReLU}\par};
                \draw[] (-1.1,-0.75+\y) rectangle (1.1,0+\y) node[pos=.5, text width=2*\backbonesscale cm,align=center] {\backbonestextreg SubM Conv $3\times 3$ stride-1\par};
                \draw[] (-1.1,-0.75+\y) rectangle (1.1,-1+\y) node[pos=.5, text width=2*\backbonesscale cm,align=center] {\backbonestexttiny \texttt{BatchNorm}, \texttt{ReLU}\par};
                \draw [decorate,decoration={brace,amplitude=3pt},xshift=-2pt,yshift=0pt] (-1.1,-1+\y) -- (-1.1,0+\y) node [black,midway,xshift=-9] {\backbonestextrepeat $\numconv{} \times$};
                \draw[] (0,-1+\y) -- (0,-1.2+\y);
                \draw[densely dotted,ultra thick,fill=black] (-\pimgsize+\idx/\stepdown,-1.2+\y) rectangle (\pimgsize-\idx/\stepdown,-2.3+\y);
                \path (-\pimgsize+\idx/\stepdown,-1.2+\y) rectangle (\pimgsize-\idx/\stepdown,-2.3+\y) node[pos=.5, text width=2*\backbonesscale cm,align=center] {\frame{\includegraphics[height=\imgheight,angle=90]{sparse/block\idx.pdf}}};
                \path (-\pimgsize+\idx/\stepdown,-1.2+\y) rectangle (\pimgsize-\idx/\stepdown,-2.3+\y) node[pos=.5, text width=2*\backbonesscale cm,align=center] {{\transparent{\filteropacity}\textcolor{black}{\rule{\filterwidth}{\filterheight}}}};
                \path (-\pimgsize+\idx/\stepdown,-1.2+\y) rectangle (\pimgsize-\idx/\stepdown,-2.3+\y) node[pos=.5, text width=2*\backbonesscale cm,align=center, text=white] {\backbonestextreg Sparse COO Pseudoimage\par};
                \draw[] (\pimgsize-\idx/\stepdown,-1.75+\y) -- (\pimgsize-\idx/\stepdown+0.2,-1.75+\y);
                \def\convtxoff{\pimgsize-\idx/\stepdown + 0.2};
                \draw[] (\convtxoff,-2.8+\y) rectangle (\convtxoff+0.5,-0.7+\y) node[pos=.5,align=center, rotate=-90] {\backbonestextreg Sparse Conv$^\mathsf{T}$\par};
                \draw[] (\convtxoff+0.5,-1.75+\y) -- (\tconvlineend,-1.75+\y);
                \ifnum\idx=1
                    \draw[] (\tconvlineend,-1.75+\y) -- (\tconvlineend,-7.45);
                \fi
                \ifnum\idx=2
                    \draw[] (\tconvlineend,-3.05+\y) rectangle (\tconvlineend+0.5,-0.45+\y) node[pos=.5,align=center, rotate=-90] {\backbonestextreg \texttt{todense}, Concat\par};
                    \draw [-stealth](\tconvlineend+0.5,-1.75+\y) -- (\tconvlineend+0.7,-1.75+\y);
                \fi
                \ifnum\idx=3
                    \draw[] (\tconvlineend,-1.75+\y) -- (\tconvlineend,-10.05);
                \fi
                
            }
         \end{tikzpicture}
         \caption{\ourmethod{}}
         \figlabel{backbonesours}
     \end{subfigure}
     \begin{tikzpicture}[overlay]
     \foreach \idx in {1, 2, 3}{
     \def\ycenter{14.13*\backbonesscale-3.5*\idx*\backbonesscale}
            \node[text width=.5cm, align=center] at (0,\ycenter) {\backbonestextblock \texttt{Conv}\\[.25em]block\\[-.75em]\idx};
        }
        \end{tikzpicture}
    \caption{PointPillars vs \ourmethod{} Backbone. The \ourmethod{} Backbone maintains and exploits pseudoimage sparsity by using SubM convs and $2 \times 2$ stride-2 convs to avoid the smearing effect of $3 \times 3$ stride-1 convs.}
    \figlabel{backbones}
\end{figure}

%% file: table_customdataset_runtimes.tex
\begin{table*}[]
\centering
\tablespacehack{}
\caption{Per instance model component runtime and standard deviation in milliseconds, run on \customdata{}'s test set, averaged over ten trials for \desktop{} and \jetsonhigh{} and averaged over three trials for \jetsonlow{} and \robot{}. Models differ only in their Feature Net and Backbone; all other components are identical. Lower is better.}
\setlength{\tabcolsep}{4.5pt}
\begin{tabular}{r|rrrrrr|r}
 & \tblheadsty{To Device} & \tblheadsty{Feature Extract} & \tblheadsty{\textbf{Feature Net}} & \tblheadsty{\textbf{Backbone}} & \tblheadsty{Head} & \tblheadsty{BBox Extract} & \tblheadsty{\textbf{Total time}} \\ \hline
\desktopdense{} & {\valsize{}0.072} {\pmsize{}$ \pm $ 0.001} & {\valsize{}1.783} {\pmsize{}$ \pm $ 0.010} & {\valsize{}0.390} {\pmsize{}$ \pm $ 0.005} & {\valsize{}2.512} {\pmsize{}$ \pm $ 0.017} & {\valsize{}0.199} {\pmsize{}$ \pm $ 0.001} & {\valsize{}15.198} {\pmsize{}$ \pm $ 0.145} & {\valsize{}20.154} {\pmsize{}$ \pm $ 0.143} \\
\desktopsparse{} & {\valsize{}0.073} {\pmsize{}$ \pm $ 0.001} & {\valsize{}1.747} {\pmsize{}$ \pm $ 0.015} & {\valsize{}0.130} {\pmsize{}$ \pm $ 0.002} & {\valsize{}6.633} {\pmsize{}$ \pm $ 0.038} & {\valsize{}0.218} {\pmsize{}$ \pm $ 0.002} & {\valsize{}4.854} {\pmsize{}$ \pm $ 0.010} & \textbf{{\valsize{}13.655} {\pmsize{}$ \pm $ 0.060}} \\ \hline
\jetsonhighdense{} & {\valsize{}0.657} {\pmsize{}$ \pm $ 0.056} & {\valsize{}13.526} {\pmsize{}$ \pm $ 0.195} & {\valsize{}4.509} {\pmsize{}$ \pm $ 0.074} & {\valsize{}17.335} {\pmsize{}$ \pm $ 0.163} & {\valsize{}1.586} {\pmsize{}$ \pm $ 0.016} & {\valsize{}126.374} {\pmsize{}$ \pm $ 0.227} & {\valsize{}163.987} {\pmsize{}$ \pm $ 0.485} \\
\jetsonhighsparse{} & {\valsize{}0.602} {\pmsize{}$ \pm $ 0.007} & {\valsize{}13.451} {\pmsize{}$ \pm $ 0.076} & {\valsize{}1.341} {\pmsize{}$ \pm $ 0.030} & {\valsize{}43.935} {\pmsize{}$ \pm $ 0.176} & {\valsize{}2.199} {\pmsize{}$ \pm $ 0.012} & {\valsize{}27.054} {\pmsize{}$ \pm $ 0.139} & \textbf{{\valsize{}88.584} {\pmsize{}$ \pm $ 0.349}} \\ \hline
\jetsonlowdense{} & {\valsize{}2.128} {\pmsize{}$ \pm $ 0.422} & {\valsize{}28.340} {\pmsize{}$ \pm $ 0.098} & {\valsize{}6.907} {\pmsize{}$ \pm $ 0.094} & {\valsize{}14.557} {\pmsize{}$ \pm $ 0.292} & {\valsize{}1.511} {\pmsize{}$ \pm $ 0.009} & {\valsize{}407.499} {\pmsize{}$ \pm $ 0.302} & {\valsize{}460.941} {\pmsize{}$ \pm $ 0.246} \\
\jetsonlowsparse{} & {\valsize{}2.163} {\pmsize{}$ \pm $ 0.106} & {\valsize{}28.813} {\pmsize{}$ \pm $ 0.047} & {\valsize{}1.728} {\pmsize{}$ \pm $ 0.007} & {\valsize{}60.233} {\pmsize{}$ \pm $ 0.076} & {\valsize{}2.385} {\pmsize{}$ \pm $ 0.004} & {\valsize{}62.169} {\pmsize{}$ \pm $ 0.073} & \textbf{{\valsize{}157.492} {\pmsize{}$ \pm $ 0.199}} \\ \hline
\robotdense{} & {\valsize{}1.531} {\pmsize{}$ \pm $ 0.216} & {\valsize{}43.073} {\pmsize{}$ \pm $ 0.482} & {\valsize{}29.237} {\pmsize{}$ \pm $ 0.319} & {\valsize{}879.225} {\pmsize{}$ \pm $ 6.116} & {\valsize{}115.363} {\pmsize{}$ \pm $ 1.065} & {\valsize{}13.706} {\pmsize{}$ \pm $ 0.064} & {\valsize{}1,082.135} {\pmsize{}$ \pm $ 6.692} \\
\robotsparse{} & {\valsize{}1.383} {\pmsize{}$ \pm $ 0.111} & {\valsize{}41.073} {\pmsize{}$ \pm $ 0.994} & {\valsize{}0.313} {\pmsize{}$ \pm $ 0.004} & {\valsize{}66.045} {\pmsize{}$ \pm $ 0.409} & {\valsize{}114.911} {\pmsize{}$ \pm $ 1.171} & {\valsize{}13.012} {\pmsize{}$ \pm $ 0.171} & \textbf{{\valsize{}236.737} {\pmsize{}$ \pm $ 2.491}}
\end{tabular}
\tablelabel{customtable}
\end{table*}

%% file: table_kitti_perf.tex
\begin{table*}[htbp]
\tablespacehack{}
\centering
\caption{Performance of PointPillars as \% AP and performance of \ourmethod{} and its ablations as the relative \% AP difference ({\tiny $\Delta$}) to PointPillars on KITTI with 16cm$\times$16cm pillars. Higher is better.
}
\setlength{\tabcolsep}{4.5pt}
\begin{tabular}{r|rrr|rrr|rrr|rrr|rrr}
        & \multicolumn{3}{c|}{\textbf{PointPillars}} 
        & \multicolumn{3}{c|}{\textbf{\ourmethod}}
        & \multicolumn{3}{c|}{\textbf{\sparseone}}
        & \multicolumn{3}{c|}{\textbf{\sparsetwo}} 
        & \multicolumn{3}{c}{\textbf{\sparsewide}} \\ 
\tblheadsty{} & Easy    & Med.    & Hard   & Easy      &  Med.     & Hard  & Easy    & Med.    & Hard   & Easy      &  Med.     & Hard  & Easy      &  Med.     & Hard  \\ \hline
\tblheadsty{BEV AP}  & 90.75    & 89.79    & 89.48   
& -5.71{\tiny $\Delta$}     & -6.53{\tiny $\Delta$}    & -4.98{\tiny $\Delta$}  
& -5.25{\tiny $\Delta$} & -5.44{\tiny $\Delta$} & -3.73{\tiny $\Delta$} 
& -5.54{\tiny $\Delta$} & -5.89{\tiny $\Delta$} & -5.60{\tiny $\Delta$} 
& -6.89{\tiny $\Delta$} & -7.24{\tiny $\Delta$} & -9.25{\tiny $\Delta$}    \\ 
\tblheadsty{3D AP}   & 82.30    & 80.34    & 79.11   
& -8.82{\tiny $\Delta$}    & -7.83{\tiny $\Delta$}    & -9.16{\tiny $\Delta$} 
& -7.62{\tiny $\Delta$} & -7.36{\tiny $\Delta$} & -8.85{\tiny $\Delta$} 
& -8.88{\tiny $\Delta$} & -7.71{\tiny $\Delta$} & -9.27{\tiny $\Delta$} 
& -11.1{\tiny $\Delta$} & -11.57{\tiny $\Delta$} & -13.69{\tiny $\Delta$}   \\ 
\end{tabular}
\tablelabel{perfbaselinesparse}
\end{table*}

%% file: table_kitti_runtimes.tex
\begin{table*}[]
\centering
\caption{Per instance model component runtime and standard deviation in milliseconds, run on KITTI's test set, averaged over ten trials. Models differ only in their Feature Net and Backbone; all other components are identical. Lower is better.}
\setlength{\tabcolsep}{4.5pt}
\begin{tabular}{r|rrrrrr|r}
 & \tblheadsty{To Device} & \tblheadsty{Feature Extract} & \tblheadsty{\textbf{Feature Net}} & \tblheadsty{\textbf{Backbone}} & \tblheadsty{Head} & \tblheadsty{BBox Extract} & \tblheadsty{\textbf{Total time}} \\ \hline
PointPillars & {\valsize{}0.064} {\pmsize{}$ \pm $ 0.000} & {\valsize{}2.375} {\pmsize{}$ \pm $ 0.010} & {\valsize{}0.303} {\pmsize{}$ \pm $ 0.004} & {\valsize{}2.358} {\pmsize{}$ \pm $ 0.027} & {\valsize{}0.204} {\pmsize{}$ \pm $ 0.002} & {\valsize{}9.105} {\pmsize{}$ \pm $ 0.030} & {\valsize{}14.410} {\pmsize{}$ \pm $ 0.033} \\
\ourmethod{} & {\valsize{}0.064} {\pmsize{}$ \pm $ 0.000} & {\valsize{}2.330} {\pmsize{}$ \pm $ 0.009} & {\valsize{}0.133} {\pmsize{}$ \pm $ 0.002} & {\valsize{}7.578} {\pmsize{}$ \pm $ 0.038} & {\valsize{}0.244} {\pmsize{}$ \pm $ 0.004} & {\valsize{}3.877} {\pmsize{}$ \pm $ 0.010} & \textbf{{\valsize{}14.226} {\pmsize{}$ \pm $ 0.056}} \\
\sparseone{} & {\valsize{}0.064} {\pmsize{}$ \pm $ 0.000} & {\valsize{}2.341} {\pmsize{}$ \pm $ 0.011} & {\valsize{}0.134} {\pmsize{}$ \pm $ 0.002} & {\valsize{}7.394} {\pmsize{}$ \pm $ 0.049} & {\valsize{}0.231} {\pmsize{}$ \pm $ 0.003} & {\valsize{}4.437} {\pmsize{}$ \pm $ 0.013} & {\valsize{}14.602} {\pmsize{}$ \pm $ 0.068} \\
\sparsetwo{} & {\valsize{}0.064} {\pmsize{}$ \pm $ 0.000} & {\valsize{}2.359} {\pmsize{}$ \pm $ 0.011} & {\valsize{}0.134} {\pmsize{}$ \pm $ 0.002} & {\valsize{}7.803} {\pmsize{}$ \pm $ 0.050} & {\valsize{}0.236} {\pmsize{}$ \pm $ 0.002} & {\valsize{}4.406} {\pmsize{}$ \pm $ 0.018} & {\valsize{}15.001} {\pmsize{}$ \pm $ 0.080} \\
\sparsewide{} & {\valsize{}0.066} {\pmsize{}$ \pm $ 0.001} & {\valsize{}2.356} {\pmsize{}$ \pm $ 0.015} & {\valsize{}0.137} {\pmsize{}$ \pm $ 0.002} & {\valsize{}17.286} {\pmsize{}$ \pm $ 0.071} & {\valsize{}0.242} {\pmsize{}$ \pm $ 0.007} & {\valsize{}6.184} {\pmsize{}$ \pm $ 0.038} & {\valsize{}26.270} {\pmsize{}$ \pm $ 0.124} 
\end{tabular}
\tablelabel{runtimes}
\end{table*}